\documentclass{article}
\usepackage{ijcai24}

\usepackage{times}
\usepackage{soul}
\usepackage{url}
\usepackage[hidelinks]{hyperref}
\usepackage[utf8]{inputenc}
\usepackage[small]{caption}
\usepackage{graphicx}
\usepackage{amsmath}
\usepackage{amsthm}
\usepackage{booktabs}
\usepackage{algorithm}
\usepackage{algorithmic}
\usepackage[switch]{lineno}
\usepackage{xcolor}
\usepackage{amssymb}
\usepackage{makecell}
\usepackage{bbding}
\usepackage{stfloats}
\usepackage{multirow}
\usepackage{CJKutf8}
\usepackage{tabularx}
\usepackage{hyperref}

\title{Vision-fused Attack: Advancing Aggressive and Stealthy Adversarial Text \\against Neural Machine Translation}

\author{
Yanni Xue$^{1}$\footnotemark[1]
\and
Haojie Hao$^{1}$\footnotemark[1]
\and
Jiakai Wang$^2$\footnotemark[2]
\and 
Qiang Sheng$^4$\and
Renshuai Tao$^5$\and
Yu Liang$^6$\and \\
Pu Feng$^1$\And
Xianglong Liu$^{1,2,3}$ 
\affiliations
$^1$State Key Laboratory of Complex \& Critical Software Environment, Beihang University, Beijing, China\\
$^2$Zhongguancun Laboratory, Beijing, China\\
$^3$Institute of Data Space, Hefei Comprehensive National Science Center, Anhui, China \\
$^4$Institute of Computing Technology, Chinese Academy of Sciences \\
$^5$Beijing Jiaotong University \\
$^6$Beijing University of Technology 
\emails
\{ynxue, haojiehao,fengpu,xlliu\}@buaa.edu.cn,
wangjk@mail.zgclab.edu.cn,
shengqiang18z@ict.ac.cn,
rstao@bjtu.edu.cn,
yuliang@bjut.edu.cn
}

\begin{document}

\maketitle

\begin{abstract}

While neural machine translation (NMT) models achieve success in our daily lives, they show vulnerability to adversarial attacks.
Despite being harmful, these attacks also offer benefits for interpreting and enhancing NMT models, thus drawing increased research attention. However, existing studies on adversarial attacks are insufficient in both attacking ability and human imperceptibility due to their sole focus on the scope of language. This paper proposes a novel vision-fused attack (VFA) framework to acquire powerful adversarial text, \emph{i.e.}, more aggressive and stealthy. 
Regarding the attacking ability, we design the vision-merged solution space enhancement strategy to enlarge the limited semantic solution space, which enables us to search for adversarial candidates with higher attacking ability. 
For human imperceptibility, we propose the perception-retained adversarial text selection strategy to align the human text-reading mechanism. Thus, the finally selected adversarial text could be more deceptive.
Extensive experiments on various models, including large language models (LLMs) like \textbf{LLaMA} and \textbf{GPT-3.5}, strongly support that VFA outperforms the comparisons by large margins (up to \textbf{81\%}/\textbf{14\%} improvements on ASR/SSIM).
\end{abstract}

\renewcommand{\thefootnote}{}

\section{Introduction}
Neural machine translation (NMT) has achieved remarkable progress and has been widely used in many scenarios with the advancement of deep neural networks. However, recent studies have revealed the vulnerability of NMT models.\footnote{Codes can be found at \href{https://github.com/Levelower/VFA}{https://github.com/Levelower/VFA}.}
A well-designed adversarial text, which aims to deceive NMT models while remaining imperceptible to humans, could result in poor model performance~\cite{zhang2021crafting}.
Nevertheless, as the old saying goes, \emph{every coin has two sides}, though harming NMT models, the adversarial attacks could also help understand the behavior of the unexplainable deep models. Therefore, generating adversarial text has essential value in constructing trustworthy and robust NMT models.

\renewcommand{\thefootnote}{\arabic{footnote}}

\setcounter{footnote}{0}

\begin{CJK*}{UTF8}{bsmi}
    \begin{figure}[t]
        \centering
        \includegraphics[width=0.48\textwidth]{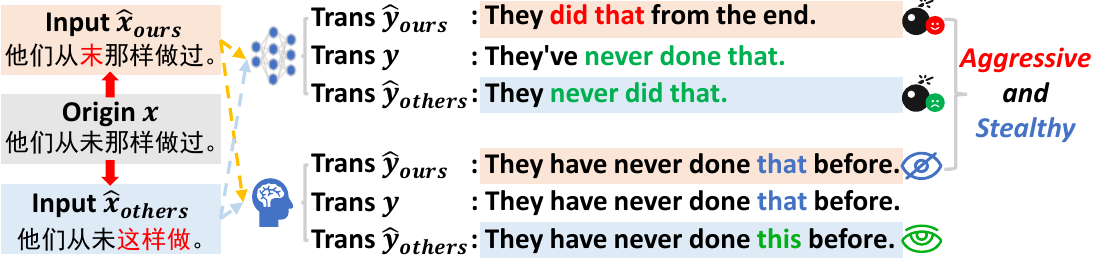}
        \caption{Our proposed VFA reduces the translation quality of NMT models by generating visually similar characters that are aggressive, such as ``未" and ``末" in the figure, while maintaining consistency with human's recognition of text to achieve stealthiness.}
        \label{fig:example}
    \end{figure}
\end{CJK*}

Existing studies on generating adversarial text for NMT models can be classified into targeted attacks and untargeted attacks based on the intended attack purpose.
Although both of them aim to fool NMT models, their purposes have a few differences. 
To be specific, the former simply misleads the translation into arbitrary wrong output, while the latter is intended to let the NMT models make particular false responses, e.g., inserting keywords into the translated sentences~\cite{2018Seq2Sick,sadrizadeh2023targeted}.

Despite significant progress in generating adversarial text, current studies still show some weaknesses, which can be summarized into two aspects:
(1) The unsatisfactory attacking ability. Most attacking methods first search candidate adversarial words in the embedding space, which is strictly constrained due to the extra semantic-preserving requirements. As a result, the searched adversarial results could not mislead the NMT models to the largest extent in such a narrow embedding space. (2) The insufficient imperceptibility goal. Current attacks mainly achieve the goal via restricting the adversarial results at the semantic level. However, the text's perception of mankind is more correlated to the visual system, i.e., humans always recognize text first and process them later. Thus, due to the sole semantic restriction, the generated adversarial text might not be stealthy enough for humans.
 
To tackle these issues, we propose Vision-fused Attack framework (VFA) to generate more aggressive adversarial text against NMT models with higher attacking ability and better visual stealthiness. Figure \ref{fig:example} illustrates the difference between our adversarial text and others.
\textbf{To improve the attacking ability}, considering that the limited semantic solution space might restrict the adversarial text candidates, we propose the vision-merged solution space enhancement (VSSE) strategy to abound the searchable adversarial candidates by the visual solution space mixture module.
In detail, we first enhance the basic semantic space with the help of a reverse translation block. Further, we map the enhanced solution space into vision space via the text-image transformation block. 
Since the mapped visual solution space mixes both semantic and visual characteristics, it offers a broader searching range for candidate adversarial words, thus making it more possible to activate higher attacking ability in practice. 
\textbf{Regarding the imperceptibility of adversarial text}, given the neglected fact that human reading accepts visual signals first to recognize text, 
we develop the perception-retained adversarial text selection (PATS) strategy to evade human perception through the perception stealthiness enhancement module.
Specifically, an improved word replacement operation is preliminarily introduced to disperse attack locations.
Then, we integrate the visual characteristics of local characters and global sentences to align with the human text-reading mechanism.
Since this selection strategy could filter the human perceptually suspected candidates, we could efficiently and accurately select more imperceptible adversarial text in principle to deceive the text perception of humans.

To demonstrate the effectiveness of the proposed method, we conduct extensive experiments under white-box and black-box settings on various representative models and widely-used datasets, including open-source and closed-source large language models like GPT-3.5 and LLaMA. The experimental results strongly support that our \textbf{VFA} outperforms the comparisons by large margins.

Our main contributions are as follows:
\begin{itemize}
    \item To the best of our knowledge, we are the first to introduce visual perception, which aligns with human reading, to generate adversarial text against NMT models.
    \item We propose a Vision-fused Attack (VFA) against NMT models, acquiring aggressive and stealthy adversarial text through vision-merged solution space enhancement and perception-retained adversarial text selection.
    \item Extensive experiments show that VFA outperforms the comparisons by large margins (up to 81\%/28\% improvements on common NMT models/LLMs), and achieves considerable imperceptibility (up to 14\% improvements) in both machine evaluation and human study.
\end{itemize}

\begin{figure*}
    \centering
    \includegraphics[width=1.0\textwidth]{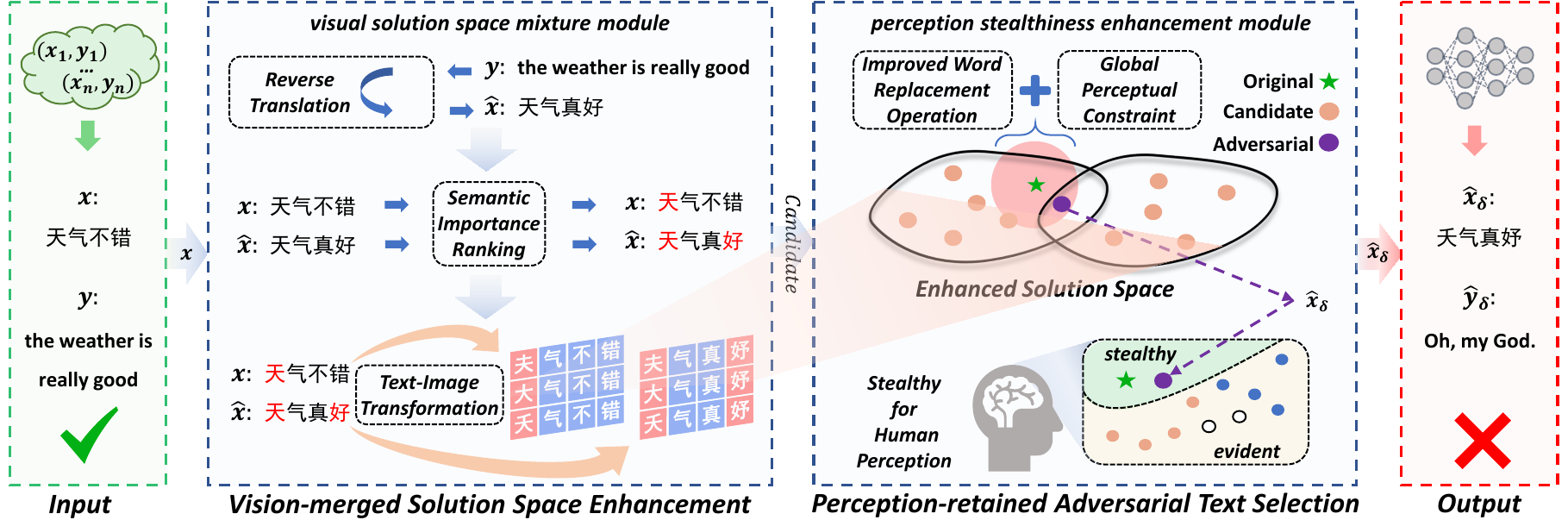}
    \caption{Overall framework of our Vision-fused Attack (VFA). We first use the Vision-merged Solution Space Enhancement strategy to search for more aggressive adversarial candidates. Then we find the final adversarial text that best matches human perception through the Perception-retained Adversarial Text Selection strategy. Finally we find more aggressive and stealthy adversarial text against NMT models.}
    \label{fig:my_image}
\end{figure*}

\section{Related Work}

The neural machine translation (NMT), which translates texts from one language to another, has achieved impressive progress using DNN models such as transformers~\cite{NIPS2017_3f5ee243}. 
Due to their excellent performance, NMT models are widely used in different applications~\cite{gao-etal-2023-improving}. 
However, recent studies of adversarial text investigate the vulnerability of NMT models, whose ultimate goal is to mislead NMT models but imperceptible to humans.
Though they are harmful to DNN models, they could also help us understand them~\cite{wang2021universal,wang2021dual}.
Therefore, there has been an increasing number of recent studies on adversarial attacks against NMT models.

Adversarial attacks on NMT models can be divided into untargeted attacks and targeted attacks. 
Untargeted attacks aim to degrade the translation quality of NMT models. Targeted attacks mislead the victim model to produce a particular translation, for example, one that does not overlap with the reference or inserts some keywords into the translation.

Current untargeted attacks always use semantic perturbations to degrade the model performance. 
Specifically, attackers rank the words in a sentence by saliency and randomly substitute them based on the saliency order~\cite{cheng-etal-2019-robust,cheng-etal-2020-advaug}.
However, random substitution may reduce the imperceptibility of adversarial text. Therefore, some studies choose to replace these words with similar ones, requiring semantic similarity techniques to identify similar word lists~\cite{feng2022language}. Nevertheless, the inefficiency of embedding techniques may still result in false substitutions. To resolve this problem, Word Saliency speedup Local Search uses Round-Trip Translation (RTT) to handle it~\cite{zhang2021crafting}. Additionally, the Doubly Round-Trip Translation (DRTT) method further amplifies the RTT to achieve better quality of adversarial text~\cite{lai2022generating}.

In targeted attack research, attacks fool the target model into generating a particular translation. Such as the translation does not overlap with the reference or push some words into it~\cite{2018Seq2Sick}.
They use a hinge-like loss term and a group lasso regularization to make perturbations, thereby achieving specific attack targets.
Since these perturbations in the embedding space are less constrained, they may not preserve semantic similarity.
Therefore, some studies define the new optimization problem, including adversarial loss and similarity terms. Finally, they perform gradient projection in the embedding space to generate adversarial sentences~\cite{sadrizadeh2023targeted}. 

Both above approaches aim to perturb original sentences with similar words, which satisfies the requirement of semantics-preserving.
However, there are differences in the cognitive process of text between humans and models. For humans, the cognitive process of text can be divided into visual perception and semantic understanding, while models only imitate the latter. Therefore, the generated adversarial text does not perform well in the visual perception part, resulting in poor human imperceptibility.

\section{Approach}

In this section, we define the adversarial text for NMT models and then elaborate on our proposed Vision-fused Attack.

\subsection{Problem Definition}

Given a source sentence $\mathbf{x}$ and a reference sentence $\mathbf{y}$, the adversarial text $\mathbf{x_{\delta}}$ is generated to mislead targeted model $\mathbb{M}$ to generate low-quality results. 
Referring to previous work~\cite{ebrahimi2018adversarial}, a successful attack satisfies:
\begin{equation}
    \begin{aligned}
         \frac{\operatorname{sim_{t}}(\mathbb{M}(\mathbf{x}),\mathbf{y})-\operatorname{sim_{t}}(\mathbb{M}(\mathbf{x_{\delta}}),\mathbf{y})}{\operatorname{sim_{t}}(\mathbb{M}(\mathbf{x}),\mathbf{y})} > \alpha.
    \end{aligned}
\end{equation}
Here, $\operatorname{sim_{t}}(\cdot)$ refers to the similarity function used for evaluating semantic similarity.
The parameter $\alpha$ is established as the lower limit, representing the lowest degradation of translation quality.
It is worth noting that we differ from previous work in assessing effectiveness, which uses sole semantic similarity~\cite{zhang2021crafting,lai2022generating}. We define an authentic adversarial text from the perspective of visual perception, and the definition of adversarial text differs accordingly.
Specifically, we generate a semantically extension $\mathbf{\hat x}$ for the original text $\mathbf{x}$ (this process will be detailed in the Reverse Translation Block), and use $\mathbf{\hat x}$ as the text to be replaced to generate the final adversarial text $\mathbf{x_{\delta}}$.
Our modified definition is given as:
\begin{equation}
\begin{aligned}
    \left\{
    \begin{aligned}
         & \frac{\operatorname{sim_{t}}(\mathbb{M}(\mathbf{x}), \mathbf{y}) - \operatorname{sim_{t}}(\mathbb{M}(\mathbf{x_\delta}), \mathbf{y})}{\operatorname{sim_{t}}(\mathbb{M}(\mathbf{x}), \mathbf{y})}>\alpha, \\
         & \operatorname{sim_{t}}(\mathbf{x}, \mathbf{\hat x})>\beta,\\
         & \operatorname{sim_{v}}(\mathbf{x}, \mathbf{x_\delta})>\theta.
    \end{aligned}
    \right.
\end{aligned}
\end{equation}
The function $\operatorname{sim_{\text{v}}}$ is a visual similarity function defined by the perception stealthiness enhancement module, which will be detailed in the following subsection. The parameter $\beta$ and $\theta$ ensure the semantic and visual similarity. The authentic adversarial text satisfies the requirements of visual and semantic similarity while degrading the translation quality.

\subsection{Overview of Vision-fused Attack}

Previous studies generated adversarial text within a limited semantic space and overlooked the significance of visual perception in text reading. Consequently, it is challenging to produce more aggressive and imperceptible results. In response, we introduce a Vision-fused Attack (VFA) to generate human imperceptible adversarial text against NMT models. The overall framework is illustrated in Figure \ref{fig:my_image}.

Addressing the limitation of the semantic solution space, we consider the amplified visual solution space and generate effective adversarial text through the Visual-merged Solution Space Enhancement (VSSE) strategy. Specifically, we expand the essential semantic solution space using a reverse translation block. Furthermore, our method generates candidate adversarial words through a text-image transformation block. With abundant adversarial candidates, we then filter unauthentic candidates using the Perception-retained Adversarial Text Selection (PATS) strategy to acquire more imperceptible adversarial text. Initially, we perform substitutions through an improved word replacement operation. Subsequently, we obtain the authentic adversarial text with a global perceptual constraint. This enables us to achieve superior perceptual retention results, aligning more closely with vision-correlated human text perception.

\subsection{Vision-merged Solution Space Enhancement}

Our vision-merged enhanced solution space consists of two parts: one is the essential semantic space expanded through reverse translation block, which is termed as $\mathbb{T}$, and the other is the mapped visual solution space corresponding to the input text, which is termed as $\mathbb{V}$.
Considering the further enhancement of vision-merged solution space, we transform the original input to increase the variety of words. Therefore, we can map semantic space into a larger visual space and search for adversarial words within this expanded area.
When generating the candidate adversarial words, we first tokenize the original input sentence and the texts which are transformed through the reverse translation block. Then, we use semantic importance ranking to obtain ordered attack locations. Finally, we acquire possible adversarial candidates through a text-image transformation block. 

\paragraph{Reverse Translation Block.} We initially expand essential semantic space through reverse translation to amplify vision-fused solution space. For the source sentence $\mathbf{x}$ and the reference sentence $\mathbf{y}$, we generate similar sentences (referred to as $\mathbf{\hat x}$) for $\mathbf{x}$.
For each reference translation $\mathbf{y}$, we use an auxiliary translation model $\operatorname{\textbf{M}_{\text{aux}}}$ to obtain transformed $\mathbf{\hat x}$ meanwhile the similarity between $\mathbf{x}$ and $\mathbf{\hat x}$ need to satisfy the constraint of the lowest thresh $\beta$. 
The sentence similarity is evaluated by a multilingual sentence model, which is termed as $\operatorname{\textbf{M}_{\text{sim}}}$:
\begin{equation}
    \begin{aligned}
        \mathbf{\hat x} = \operatorname{\textbf{M}_{\text{aux}}}(\mathbf{x}), ~ \operatorname{\textbf{M}_{\text{sim}}}(\mathbf{\hat x},\mathbf{x}) > \beta.
    \end{aligned}
\end{equation}

\paragraph{Semantic Importance Ranking.} The source sentence can be represented as a list of words $\mathbf{ w} = \langle w_1, w_2, ..., w_n \rangle$. For a sequence that is masked at position $i$, which is termed as $\mathbf{w^i_{\text{mask}}} = \langle w_1, w_2, ..., w_{i-1}, [m], w_{i+1}, ..., w_n \rangle$. We calculate the importance scores of these words at different positions named $\mathbf{w}^i_{{\text{imp}}}$. Masked language model $\operatorname{\textbf{M}_{\text{mlm}}}$ (e.g., BERT) is used to predict word probability of occurrence. This process can be written as:
\begin{equation}
\begin{aligned}
   \mathbf{w}^i_{{\text{imp}}} = \operatorname{\textbf{M}_{\text{mlm}}}(w_i|\langle w_1, ..., w_{i-1}, [m], w_{i+1}, ..., w_n \rangle).
\end{aligned}
\end{equation}

\paragraph{Text-image Transformation Block.} For each character $c$ in the Unicode character dictionary
(referred to as $\mathbb{C}$), we conduct a level-wise search to find the most similar candidate adversarial characters. Initially, we follow previous work to utilize the glyph dictionary, denoted as $\mathbb{D}$, which stores mappings of specific characters $c$ and their radicals $z$ (Here we denote the set of all radicals as $\mathbb{Z}$, where $c\in \mathbb{C}$ and $z\in \mathbb{Z}$)~\cite{su2022rocbert}. Each radical of character $c$ can be found through function $f_c(\cdot)$. We start by aggregating characters with the same radical (denoted as $c'$), narrowing down the candidate pool. We can obtain a portion of the candidate set $\textbf{S}_{\text{rad}}$ through a similar radical search, denoted as $f_a(\cdot)$:
\begin{equation}
\begin{aligned}
\left\{
    \begin{aligned}
& f_c:c \to \{ z \, | \, z \in \mathbb{Z} ,(c,z) \in \mathbb{D} \}, \\
& f_a:c \to \{ c' \, | \, f_c(c) \cap f_c(c') \neq \emptyset \}.
 \end{aligned}
    \right.
\end{aligned}
\end{equation}
Due to the limitations of the glyph dictionary, we further pixelated the characters set and conducted similar image searches to find similar characters, which is denoted as $\textbf{S}_{\text{pix}}$. The function $\text{p}(\cdot)$ is defined to convert a sentence or character into the correlated image. We map the input character to its pixelated image through $\text{p}(c)$.
Then, we calculate cosine similarity to search $top$ $m$ visually similar results using Faiss (a tool that accelerates vector calculations through hierarchical search)~\cite{johnson2019billion}. And the procedure could be formulated as $f_{cos}(\cdot)$:
\begin{equation}
\begin{aligned}
f_{\text{cos}}:c  \to \{ top(m, cos(\text{p}(c), \text{p}(c'))) ,  c' \in \mathbb{C}\},
\end{aligned}
\end{equation}
where the $top(\cdot)$ represents the function that identifies the highest-ranked elements based on certain scores, such as cosine similarity.
Finally, we apply Mean Squared Error (MSE) similarity to re-rank the possible adversarial character and select the $top$ $k$ candidate result:
\begin{equation}
\begin{aligned}
\mathbf{S}_{\text{pix}} = top(k, \text{mse}(f_{\text{cos}}(c),c)).
\end{aligned}
\end{equation}
Combined with the results of $\textbf{S}_{\text{rad}}$ and $\textbf{S}_{\text{pix}}$ , we can obtain the final candidate adversarial characters set $\textbf{S}$.
\begin{equation}
\begin{aligned}
 \textbf{S} = \textbf{S}_{\text{rad}} \cup \textbf{S}_{\text{pix}}.
\end{aligned}
\end{equation}

\subsection{Perception-retained Adversarial Text Selection}

For the generated candidate characters set $\textbf{S}$, we apply a perception stealthiness enhancement module, which consists of improved word replacement operation and global perceptual constraint, to filter out truly effective adversarial text. Therefore, the visually imperceptible candidates could stand out to better mislead human perception, \emph{i.e.}, stealthy.

\paragraph{Improved Word Replacement Operation.} Firstly, we implement a substitution constraint strategy grounded in human perception. We intuitively regulate the replacement rate, deliberately replacing only one character within each word. This strategy disrupts the semantic expression of the word and minimizes the impact of perturbation.
For text $\mathbf{\hat x}$ to be replaced, 
$w_i$ denotes the $i$-th word in order of importance $\mathbf{w}^i_{{\text{imp}}}$ of the text, and $c_j$ signifies the $j$-th character of a word $w_i$. The replacement operation rate, denoted as $r$, falls within the range $0 \leq r \leq 1$, representing the replacement probability of the overall sentence. And function $\operatorname{rep}(c_{j_\delta}, c_j)$ indicates whether the character $c_j$ is substituted by $c_{j_\delta}$ in $\mathbf{S}$. When the character changes, the function equals 1. Otherwise, it equals 0. We can detail this constraint as follows:
\begin{equation}
\begin{aligned}
\left\{
    \begin{aligned}
\frac{\sum_{w_i \in \mathbf{x_\delta}} \sum_{c_j \in w_i} \operatorname{rep}(c_{j_\delta}, c_j)}{\sum_{w_i \in \mathbf{\hat x}} \sum_{c_j \in w_i} 1} < r,  & \\
   \forall w_i \in \mathbf{x_\delta}, \quad \sum_{c_j \in w_i} \operatorname{rep}(c_{j_\delta}, c_j) \leq 1. &
    \end{aligned}
    \right.
\end{aligned}
\end{equation}

\paragraph{Global Perceptual Constraint.} Moreover, we take visual perception constraints into account. We introduce a visual perceptual similarity score for batch assessment of visual similarity.
For a source sentence $\mathbf{x}$ paired with a reference image represented as $\text{p}(\mathbf{x})$, and a perturbed sentence $\mathbf{x_{\delta}}$ with its associated image denoted as $\text{p}(\mathbf{x_{\delta}})$, $\mathbb{L}(a, b)$ denotes the perceptual similarity score between images $a$ and $b$. The parameter $\epsilon$ serves as a weight for local perception, where $0 \leq \epsilon \leq 1$.
The sentence visual similarity score can be calculated using the visual perception constraint strategy. We introduce the LPIPS metric to construct a global perceptual constraint~\cite{zhang2018unreasonable}. This constraint measures the global perceptual similarity of a sentence and aggregates the local perceptual similarities of characters with the weighted summation. Finally, we use the perceptual constraint threshold $\theta$ to constrain visual similarity:
\begin{equation}
\begin{aligned}
    \mathbb{L}(\text{p}(\mathbf{x_\delta}), \text{p}(\mathbf{x})) + \epsilon\sum_{w_i \in \mathbf{x_\delta}} \sum_{c_j \in w_i} \mathbb{L}(\text{p}(c_{j_\delta}), \text{p}(c_j)) > \theta.
\end{aligned}
\end{equation}

\begin{table*}[htbp]
    \footnotesize
    \centering
    \tabcolsep=0.1cm
    \renewcommand{\arraystretch}{0.95}
    \begin{tabular*}{\hsize}{@{}@{\extracolsep{\fill}}llcclcclcclcc@{}}
        \toprule
            \multirow{2}{*}{Method} & \multicolumn{3}{c}{WMT19 (Zh-En)}& \multicolumn{3}{c}{WMT18 (Zh-En)} & \multicolumn{3}{c}{TED (Zh-En)} & \multicolumn{3}{c}{ASPEC (Ja-En)} \\
            \cmidrule(lr){2-4} \cmidrule(lr){5-7} \cmidrule(lr){8-10} \cmidrule(lr){11-13}
            & BLEU{\color{blue} ↓} &   ASR{\color{red} ↑} & SSIM{\color{red} ↑} & BLEU{\color{blue} ↓}   & ASR{\color{red} ↑} & SSIM{\color{red} ↑}  & BLEU{\color{blue} ↓}   & ASR{\color{red} ↑} & SSIM{\color{red} ↑} & BLEU{\color{blue} ↓}   & ASR{\color{red} ↑} & SSIM{\color{red} ↑} \\
        \midrule
            Baseline&$0.178$&&&$0.163$&&&$0.159$&&&$0.075$&& \\
            HotFlip&$0.141_{\downarrow 21\%}$&$0.213$&$0.717$&$0.131_{\downarrow 19\%}$&$0.212$&$0.722$&$0.121_{\downarrow24\%}$&$0.208$&$0.737$&$0.047_{\downarrow 38\%}$&$0.334$&$0.717$ \\
            Seq2Sick&$0.139_{\downarrow 22\%}$&$0.198$&$0.773$&$0.134_{\downarrow 18\%}$&$0.164$&$0.777$&$0.112_{\downarrow 30\%}$&$0.228$&$0.762$&$0.072_{\downarrow 4\%}$&$0.030$&$\mathbf{0.862}$ \\
            Targeted&$0.134_{\downarrow 25\%}$&$0.234$&$0.799$&$0.126_{\downarrow 22\%}$&$0.211$&$0.801$&$0.114_{\downarrow 28\%}$&$0.266$&$0.793$&$0.047_{ \downarrow37\%}$&$0.308$&$0.734$ \\
            DRTT&$0.144_{\downarrow 19\%}$&$0.173$&$0.768$&$0.131_{\downarrow 19\%}$&$0.173$&$0.760$&$0.136_{\downarrow14\%}$&$0.130$&$0.780$&$0.069_{\downarrow 8\%}$&$0.063$&$0.850$ \\
            ADV&$0.146_{\downarrow 18\%}$&$0.174$&$0.838$&$0.142_{\downarrow 12\%}$&$0.141$&$0.843$&$0.125_{\downarrow 21\%}$&$0.200$&$0.842$&\multicolumn{1}{c}{-}&-&- \\
            \textbf{Ours}&$\mathbf{0.107}_{\downarrow\mathbf{40\%}}$&$\mathbf{0.382}$&$\mathbf{0.950}$&$\mathbf{0.097}_{\downarrow\mathbf{40\%}}$&$\mathbf{0.384}$&$\mathbf{0.949}$&$\mathbf{0.109}_{\downarrow\mathbf{31\%}}$&$\mathbf{0.299}$&$\mathbf{0.964}$&$\mathbf{0.042}_{\downarrow\mathbf{44\%}}$&$\mathbf{0.387}$&$0.859$ \\
        \bottomrule
    \end{tabular*}
    \caption{Performance of VFA on Zh-En and Ja-En translation tasks using pure-text NMT models. The "Baseline" row records the metric scores on the clean dataset. The remaining rows record the different metric scores and the decrease relative to the “Baseline” of adversarial texts generated by various methods. \textcolor{blue}{$\downarrow$} indicates the lower, the better, and \textcolor{red}{$\uparrow$} is the opposite.}
    \label{tab:main_table}
\end{table*}

\section{Experiment}

In this section, we first describe the experimental settings, and then we report the experimental results and some discussions on the common NMT models and LLMs.

\subsection{Experiments Settings }

\paragraph{Datasets and Models.} We choose the validation set of WMT19~\cite{ng-etal-2019-facebook}, WMT18~\cite{bojar-etal-2018-findings}, and TED~\cite{cettolo-etal-2016-iwslt} for the Chinese-English (Zh-En) translation task and the test set of ASPEC~\cite{NAKAZAWA16.621} for the Japanese-English (Ja-En) translation task. Regarding the models, the NMT models for both translation tasks are implemented using HuggingFace's Marian Model~\cite{mariannmt}, with the Zh-En/Ja-En translation models as the targeted models and the En-Zh/En-Ja models as the auxiliary models. These datasets and models are widely used in previous studies. Besides, we consider pixel-based machine translation model as the targeted model to test the validity of our method~\cite{salesky2023multilingual}. 

\paragraph{Evaluation Metrics.} We use the relative decrease of the BLEU to measure the aggressiveness of adversarial text~\cite{papineni-etal-2002-bleu}. A successful attack is defined when the BLEU score decreases by over 50\%. The attack success rate (ASR) is defined as the ratio of successful adversarial texts to the total. Finally, we use the SSIM value to evaluate the imperceptibility of the adversarial text~\cite{SSIM}.

\paragraph{Compared Methods.} We choose several state-of-the-art works about NLP attack and NMT attack, including HotFlip~\cite{ebrahimi2018hotflip}, Seq2Sick~\cite{2018Seq2Sick}, Targeted Attack~\cite{sadrizadeh2023targeted}, DRTT~\cite{lai2022generating} and ADV~\cite{su2022rocbert}.

\paragraph{Implementation Details.} As for the hyperparameter settings, we set the global perception constraint to 0.95 and the replacement rate to 0.2. To evaluate the semantic similarity between two sentences, we employ the HuggingFace sentence-transformer model~\cite{wang2020minilm}, which supports multiple languages.
Additionally, we utilize a Bert architecture model~\cite{chinese-bert-wwm} to predict the importance of words.
We conduct experiments in a cluster of NVIDIA GeForce RTX 3090 GPUs. 

\subsection{Effectiveness on Common NMT Models}

In this section, we evaluate the aggressiveness and imperceptibility of adversarial texts generated by our VFA and compared methods.
The evaluation is conducted on the Zh-En and Ja-En translation tasks using pure-text and pixel-based NMT models. 

As seen in Tables \ref{tab:main_table} and \ref{tab:pixel}, \textbf{our VFA generates adversarial texts with the highest aggressiveness across both pixel-based and pure-text models in both tasks.}

(1) In terms of aggressiveness, our VFA achieves the maximum BLEU decrease and the highest ASR on different datasets and translation tasks. Taking the results on WMT18 as an example, our VFA achieves a BLEU decrease of 40\%, better than the best of 22\% achieved by the Targeted Attack. Our ASR (0.384) outperforms the best (0.212) by 81\%. This indicates that our proposed Vision-merged Solution Space Enhancement strategy effectively improves aggressiveness.

(2) Regarding imperceptibility, our VFA also achieves the best result. Our VFA maintains an SSIM value above 0.94 on Zh-En translation tasks, while the ADV has an SSIM value of no more than 0.85. Although our VFA is second only to Seq2Sick in the imperceptibility of the Ja-En translation task, Seq2Sick nearly keeps the original sentence when attacking Japanese texts, resulting in lower ASR and inability to guarantee aggressiveness. The results demonstrate that our proposed Perception-retained Adversarial Text Selection strategy effectively improves the imperceptibility of adversarial texts.

\begin{table}[t]
    \tabcolsep=0.03cm
    \footnotesize
    \centering
    \begin{tabular}{llclclc}
        \toprule
            \multirow{2}{*}{Method} & \multicolumn{2}{c}{WMT19 (Zh-En)}& \multicolumn{2}{c}{WMT18 (Zh-En)} & \multicolumn{2}{c}{TED (Zh-En)} \\
            \cmidrule(lr){2-3} \cmidrule(lr){4-5} \cmidrule(lr){6-7}
            & BLEU{\color{blue} ↓} & ASR{\color{red} ↑} & BLEU{\color{blue} ↓}   & ASR{\color{red} ↑}  & BLEU{\color{blue} ↓}   & ASR{\color{red} ↑} \\
        \midrule
            Baseline&$0.069$&&$0.070$&&$0.165$& \\
            HotFlip&$0.054_{\downarrow 22\%}$&$0.216$&$0.057_{\downarrow 18\%}$&$0.201$&$0.130_{\downarrow 21\%}$&$0.202$ \\
            Seq2Sick&$0.055_{\downarrow 21\%}$&$0.209$&$0.058_{\downarrow 17\%}$&$0.189$&$0.134_{\downarrow 18\%}$&$0.183$ \\
            Targeted&$0.056_{\downarrow 19\%}$&$0.162$&$0.060_{\downarrow 15\%}$&$0.154$&$0.143_{\downarrow 13\%}$&$0.105$ \\
            DRTT&$0.058_{\downarrow 16\%}$&$0.139$&$0.060_{\downarrow 14\%}$&$0.125$&$0.140_{\downarrow 15\%}$&$0.116$ \\
            ADV&$0.062_{\downarrow 10\%}$&$0.133$&$0.063_{\downarrow 10\%}$&$0.123$&$0.138_{\downarrow 16\%}$&$0.149$ \\
            \textbf{Ours}&$\mathbf{0.053}_{\downarrow\mathbf{23\%}}$&$\mathbf{0.232}$&$\mathbf{0.056}_{\downarrow\mathbf{20\%}}$&$\mathbf{0.220}$&$\mathbf{0.128}_{\downarrow\mathbf{23\%}}$&$\mathbf{0.215}$ \\
        \bottomrule
    \end{tabular}
    \caption{Performance of VFA on the pixel-based NMT model.}
    \label{tab:pixel}
\end{table}

\begin{table*}[htbp]
    \footnotesize
    \centering
    \tabcolsep=0.1cm
    \renewcommand{\arraystretch}{0.95}
    \begin{tabular*}{\hsize}{@{}@{\extracolsep{\fill}}lclclclcclc@{}}
        \toprule
            \multirow{2}{*}{Method} & \multirow{2}{*}{Model} & \multicolumn{2}{c}{WMT19}& \multicolumn{2}{c}{WMT18} & \multicolumn{2}{c}{TED} & \multirow{2}{*}{Model} &\multicolumn{2}{c}{WMT19} \\
            \cmidrule(lr){3-4} \cmidrule(lr){5-6} \cmidrule(lr){7-8} \cmidrule(lr){10-11}
            &  & BLEU{\color{blue} ↓} & ASR{\color{red} ↑} & BLEU{\color{blue} ↓} & ASR{\color{red} ↑} & BLEU{\color{blue} ↓}  & ASR{\color{red} ↑} & & BLEU{\color{blue} ↓} &   ASR{\color{red} ↑} \\
        \midrule
            Baseline&\multirow{7}{*}{\makecell{LLaMA \\ (open)}}&$0.156$&&$0.146$&&$0.113$&&\multirow{7}{*}{\makecell{ChatGPT \\ (closed)}}&$0.226$& \\
            HotFlip&&$0.119_{\downarrow 23\%}$&$0.289$&$0.113_{\downarrow 23\%}$&$0.249$&$0.083_{\downarrow 27\%}$&$0.273$&&$0.188_{\downarrow 17\%}$&$0.205$ \\
            Seq2Sick&&$0.118_{\downarrow 24\%}$&$0.273$&$0.113_{\downarrow 23\%}$&$0.241$&$0.073_{\downarrow 35\%}$&$0.307$&&$0.180_{\downarrow 20\%}$&$0.235$ \\
            Targeted&&$0.115_{\downarrow 26\%}$&$0.306$&$0.109_{\downarrow 25\%}$&$0.281$&$0.081_{\downarrow 29\%}$&$0.298$&&$0.175_{\downarrow 23\%}$&$0.237$ \\
            DRTT&&$0.126_{\downarrow 19\%}$&$0.233$&$0.119_{\downarrow 18\%}$&$0.224$&$0.098_{\downarrow 14\%}$&$0.166$&&$0.181_{\downarrow 20\%}$&$0.231$ \\
            ADV&&$0.132_{\downarrow 15\%}$&$0.214$&$0.128_{\downarrow 12\%}$&$0.174$&$0.093_{\downarrow 18\%}$&$0.213$&&$0.200_{\downarrow 12\%}$&$0.156$ \\
            Ours&&$\mathbf{0.099}_{\downarrow\mathbf{37\%}}$&$\mathbf{0.385}$&$\mathbf{0.094}_{\downarrow\mathbf{35\%}}$&$\mathbf{0.359}$&$\mathbf{0.071}_{\downarrow\mathbf{37\%}}$&$\mathbf{0.325}$&&$\mathbf{0.168}_{\downarrow\mathbf{26\%}}$&$\mathbf{0.281}$ \\
        \midrule
            Baseline&\multirow{7}{*}{\makecell{BaiChuan \\ (open)}}&$0.227$&&$0.199$&&$0.145$&&\multirow{7}{*}{\makecell{ERNIE \\ (closed)}}&$0.275$& \\
            HotFlip&&$0.187_{\downarrow 17\%}$&$0.182$&$0.167_{\downarrow 17\%}$&$0.155$&$0.113_{\downarrow 22\%}$&$0.201$&&$0.223_{\downarrow 19\%}$&$0.157$ \\
            Seq2Sick&&$0.185_{\downarrow 18\%}$&$0.169$&$0.167_{\downarrow 17\%}$&$0.171$&$\mathbf{0.105}_{\downarrow\mathbf{27\%}}$&$0.215$&&$0.213_{\downarrow 22\%}$&$0.198$ \\
            Targeted&&$0.179_{\downarrow 21\%}$&$0.199$&$0.159_{\downarrow 20\%}$&$0.203$&$0.107_{\downarrow 26\%}$&$\mathbf{0.228}$&&$0.209_{\downarrow 24\%}$&$0.213$ \\
            DRTT&&$0.182_{\downarrow 20\%}$&$0.183$&$0.161_{\downarrow 19\%}$&$0.199$&$0.126_{\downarrow 13\%}$&$0.133$&&$0.220_{\downarrow 20\%}$&$0.187$ \\
            ADV&&$0.206_{\downarrow 9\%}$&$0.110$&$0.183_{\downarrow 8\%}$&$0.110$&$0.122_{\downarrow 16\%}$&$0.143$&&$0.244_{\downarrow 11\%}$&$0.113$ \\
            Ours&&$\mathbf{0.176}_{\downarrow\mathbf{23\%}}$&$\mathbf{0.211}$&$\mathbf{0.158}_{\downarrow\mathbf{21\%}}$&$\mathbf{0.209}$&$0.110_{\downarrow 24\%}$&$0.209$&&$\mathbf{0.205}_{\downarrow\mathbf{25\%}}$&$\mathbf{0.223}$ \\
        \bottomrule
    \end{tabular*}
    \caption{Performance of our VFA on open-source and closed-source LLMs, respectively. Our VFA achieves considerable attacking ability.}
    \label{tab:LLM_table}
\end{table*}

\subsection{Transferability on LLMs}

In this section, we evaluate the transferability of our VFA through LLM testing.
We use adversarial texts generated by the common NMT model to evaluate the BLEU decrease and ASR on LLM. Our selection includes four models: LLaMA-13B~\cite{touvron2023llama}, BaiChuan-13B~\cite{baichuan2023baichuan2}, GPT-3.5-turbo~\cite{ChatGPT}, and Wenxin Yiyan (ERNIE)~\cite{ERNIE}, representing leading models in both open-source and closed-source fileds. ChatGPT and LLaMA represent the most advanced LLM for English, while ERNIE and BaiChuan represent the most advanced LLM for Chinese. These LLMs perform much better on clean datasets than common NMT models, indicating that LLMs also have strong capabilities in translation tasks.

Table \ref{tab:LLM_table} displays the experiments on these models. These results indicate that \textbf{even large language models exhibit decreased performance in the face of attack.} We further give some insights and discussions as follows:

(1) Our VFA demonstrates the best aggressiveness in the open-source and closed-source LLMs. For the widely used LLaMA and ChatGPT, our VFA achieves the strongest aggressiveness on all datasets. For BaiChuan and ERNIE, which perform better in Chinese, our VFA also performs best on WMT19 and WMT18. This proves that our VFA has good transferability for LLMs.

(2) An interesting phenomenon is that our VFA is more aggressive against English LLMs than other methods. However, when applied to Chinese LLMs, our VFA only achieved a slight lead in aggressiveness and even performed weaker than Targeted Attack and Seq2Sick on the TED dataset. We attribute this phenomenon to the learning of human perception by LLM. After training with a large amount of Chinese corpus, the large model can generalize to a certain degree of human-like perception ability, which makes them robust to visual adversarial texts generated by our VFA.

\subsection{Human Study}

To evaluate the impact of adversarial texts generated by our VFA and compared methods on reading comprehension, we conduct a human perception study on \href{https://www.surveyplus.cn/}{SurveyPlus}, which is one of the most commonly used crowdsourcing platforms. 

We select 20 adversarial texts generated by six methods that meet the definition of a successful attack, then we select 105 subjects and conduct the following human perception experiments: 
(1) \textbf{Semantic understanding.} Subjects are informed of some changes in each text and asked if these changes affect their understanding of semantics. 
(2) \textbf{Semantic comparison.} For the texts selected by the subjects in the previous stage, they are required to compare them with original texts and choose the ones that maintain semantic consistency. 
Finally, we count the number of texts in each method that do not affect understanding (score1) and are semantically consistent with the original texts (score2), and divide them by the total number of people as the scores for each method.

From Table \ref{tab:human_evaluation}, it can be seen that our VFA achieves the second-highest score in the semantic understanding stage and the highest score in the semantic comparison stage. This indicates that our method has the least impact on human understanding and the least impact on semantic changes. As for DRTT, which achieved the highest score in semantic understanding, due to its use of grammatically similar modification strategies, it will have a significant change in the semantics of the original sentence, resulting in a lower score in the semantic comparison stage. 
In the total score of the two stages, VFA achieves the highest, indicating that it has the least impact on human reading comprehension and has good imperceptibility.

\begin{table}[t]
    \footnotesize
    \centering
    \tabcolsep=0.1cm
    \renewcommand{\arraystretch}{0.9}
    \begin{tabular*}{\hsize}{@{}@{\extracolsep{\fill}}cccccccc@{}}
        \toprule
            Method & Ours & ADV & DRTT & HotFlip & Targeted & Seq2Sick   \\
        \midrule
            score1{\color{red} ↑}&$9.35$&$6.13$&$\mathbf{9.76}$&$5.07$&$6.66$&$5.90$ \\
            score2{\color{red} ↑}&$\mathbf{8.63}$&$5.77$&$5.49$&$4.31$&$4.29$&$3.88$ \\
            sum{\color{red} ↑}&$\mathbf{17.98}$&$11.90$&$15.25$&$9.38$&$10.95$&$9.78$ \\
        \bottomrule
    \end{tabular*}
    \caption{Result of human study. We add up the scores of the two stages (score1+score2) as the total score (sum) for each method.}
    \label{tab:human_evaluation}
\end{table}

\subsection{Ablation Study}

In this section, we conducted several ablation studies to further investigate the contributions of some crucial components and hyper-parameters in our method. 

\begin{table}[b]
    \footnotesize
    \centering
    \tabcolsep=0.05cm
    \renewcommand{\arraystretch}{0.9}
    \begin{tabular*}{\hsize}{@{}@{\extracolsep{\fill}}cccclcc@{}}
        \toprule
            Ablation & VSSE & PATS & TIT & BLEU{\color{blue} ↓} & ASR{\color{red} ↑} & SSIM{\color{red} ↑} \\
        \midrule
            \multirow{4}{*}{\makecell{VSSE \\ + \\ PATS}}     & \Checkmark   & \Checkmark   & rad+pix   & $0.107_{\downarrow 40\%}$   & $0.382$ & $0.950$ \\
                                                          & \Checkmark   & \XSolidBrush & rad+pix   & $0.077_{\downarrow 57\%}$   & $0.542$ & $0.922$ \\
                                                          & \XSolidBrush & \Checkmark   & rad+pix   & $0.176_{ \downarrow 0.8\%}$ & $0.013$ & $0.993$ \\
                                                          & \XSolidBrush & \XSolidBrush & rad+pix   & $0.117_{\downarrow 34\%}$   & $0.316$ & $0.662$ \\
        \midrule
            \multirow{3}{*}{TIT}                           & \Checkmark   & \Checkmark   & rad+pix   & $0.107_{\downarrow 40\%}$   & $0.382$ & $0.950$ \\
                                                          & \Checkmark   & \Checkmark   & pix     & $0.092_{\downarrow 49\%}$   & $0.476$ & $0.944$ \\
                                                          & \Checkmark   & \Checkmark   & rad     & $0.125_{ \downarrow 30\%}$  & $0.283$ & $0.955$ \\
        \bottomrule
    \end{tabular*}
    \caption{Effectiveness of different components in our VFA.}
    \label{tab:ablation_for_component}
    
\end{table}

\paragraph{Effectiveness of Enhanced Solution Space and Perception Constraint.} We divide our VFA into Vision-merged Solution Space Enhancement (VSSE) and Perception-retained Adversarial Text Selection (PATS). For the vision-merged solution space, we compare the aggressiveness with the semantic solution space. Additionally, we explore the impact of PATS on the imperceptibility of adversarial texts.
We conduct four sets of ablation experiments on these two components, namely, whether to use VSSE and PATS. As shown in Table \ref{tab:ablation_for_component}, it is evident that searching in the enhanced solution space significantly improves aggressiveness compared to semantic solution space. Additionally, the PATS ensures a higher SSIM similarity score between the original texts and adversarial texts, which means ensuring the imperceptibility of adversarial texts. Meanwhile, adversarial texts generated using only semantic solution space violate the visual imperceptible rule. Therefore, after applying PATS, the ASR of semantic space search will significantly decrease.

\begin{figure}[t]
    \centering
    \includegraphics[width=0.48\textwidth]{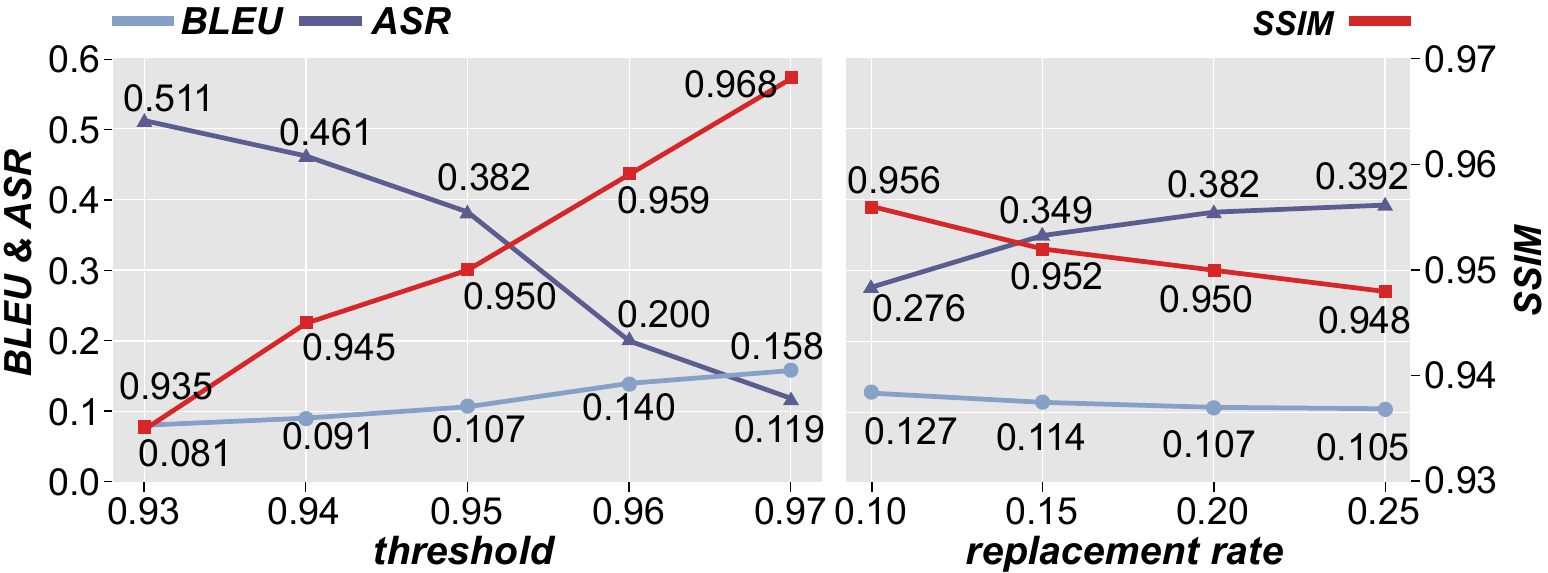}
    \caption{Effectiveness of hyper-parameters.}
    \label{fig:ablation-parameter}
\end{figure}

\paragraph{Effectiveness of Different Parts in Text-Image Transformation.} In the Text-image Transformation (TIT) block, we employed two strategies to form a complete visual solution space, one formed through similar radical components (rad) and the other through pixel characters (pix). Therefore, we explore the impact of these two complementary strategies on aggressiveness and imperceptibility.
As shown in Table \ref{tab:ablation_for_component}, the results indicate that the visual solution space formed by pixel-based strategies has stronger aggressiveness, while radical-based strategies have stronger imperceptibility. Therefore, by using the former as a supplement to the latter, we ultimately form a solution space that is both aggressive and has extremely high imperceptibility.

\paragraph{Impacts of Hyper-Parameters.} We analyze the impact of various visual perception constraint thresholds $\theta$ (threshold) and different replacement rates $r$ (rate) of aggressiveness and imperceptibility. As shown in Figure \ref{fig:ablation-parameter}, we can draw the following conclusion: 
(1) It can be observed that the stronger the visual perception constraint, the higher the SSIM, indicating that visual perception constraint can significantly improve the imperceptibility of adversarial texts.
(2) It can be observed that a higher replacement rate leads to an increase in ASR but a decrease in SSIM, which is consistent with our expectations. As more words are replaced, the difference between the adversarial texts and the original texts becomes greater, and the aggressiveness also increases. 
We balanced aggressiveness and imperceptibility by combining the replacement rate and visual perception constraints.

\begin{figure}[ht]
    \centering
    \includegraphics[width=0.48\textwidth]{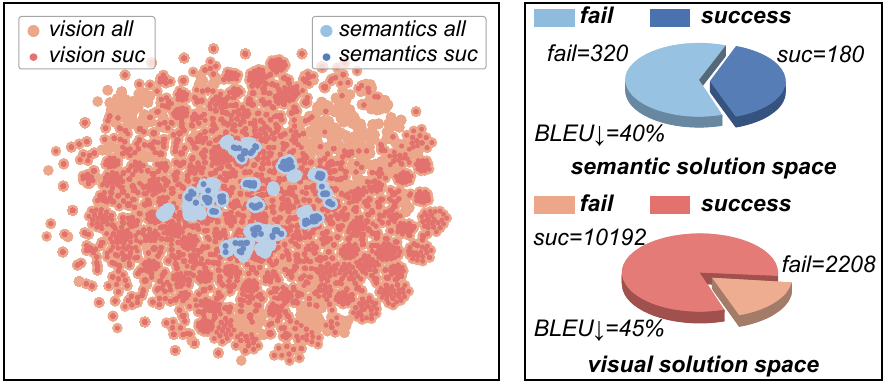}
    \caption{Enhanced solution space and semantic solution space.}
    \label{fig:analysis-space}
\end{figure}
\subsection{Case Study}

\paragraph{Enhanced Solution Space Analysis.} We analyze the enhanced solution space from a visualization perspective. For a single text, we calculated embeddings for all texts in the enhanced solution space and the semantic solution space, then reduced the dimensionality using t-SNE~\cite{JMLR:v9:vandermaaten08a}. The result is shown in Figure \ref{fig:analysis-space}. 
The left side of the figure shows the sample distribution in the two solution spaces, while the right side indicates the number of successfully attacked texts and the corresponding BLEU decrease in both spaces. Generally, we can draw such conclusions: 
(1) It can be witnessed that the scope of the enhanced solution space is broader than that of the semantic solution space. This indicates that the enhanced solution space can obtain more diverse adversarial texts, providing more possibilities to find more effective adversarial texts. 
(2) The pie charts of both solution spaces show that the enhanced solution space generates more aggressive texts than the semantic solution space. This is further proven by the BLEU decrease of the two approaches. Therefore, an enhanced solution space helps generate more texts with stronger aggressiveness.

\begin{figure}[t]
    \centering
    \includegraphics[width=0.48\textwidth]{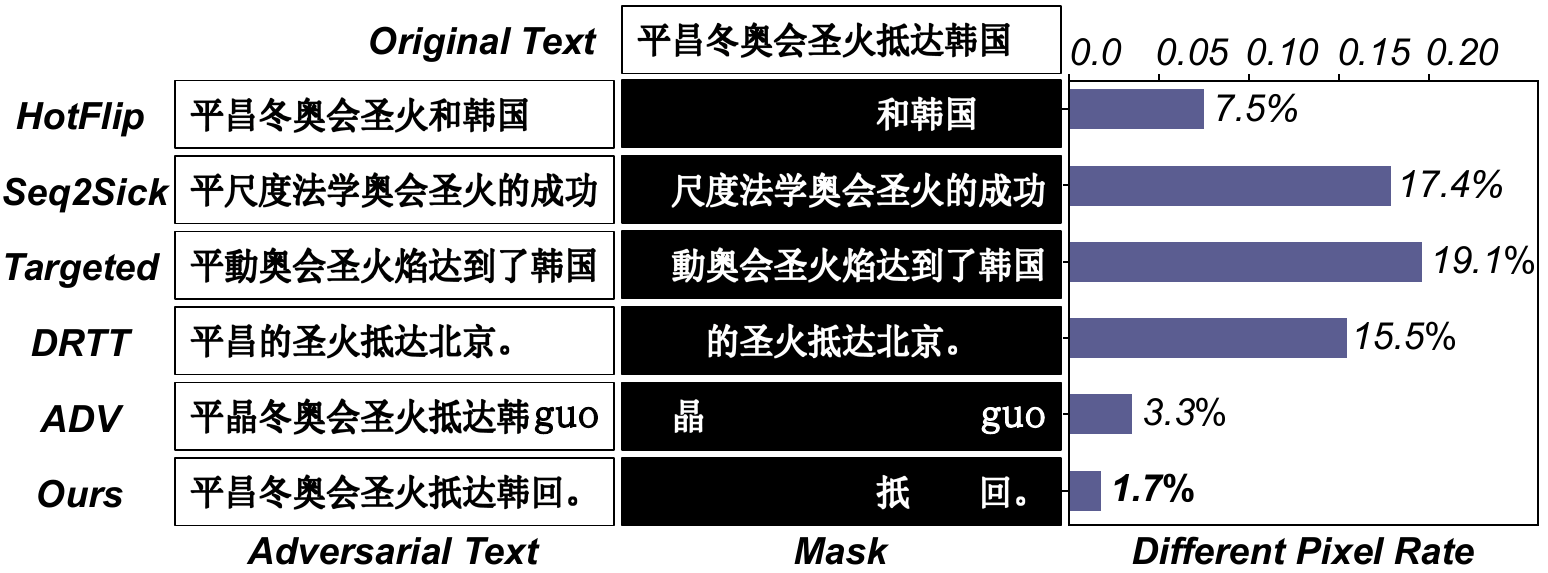}
    \caption{Adversarial texts of all methods and their masks.}
    \label{fig:analysis-imperceptibility}
\end{figure}

\paragraph{Imperceptibility Analysis.} We visualized the adversarial texts generated by all methods and analyzed the imperceptibility of individual examples. 
Figure \ref{fig:analysis-imperceptibility} shows the adversarial texts generated by our method and five compared methods. The differences between them and the original texts are presented in the form of masks. The right side of the figure indicates the proportion of the differences in pixel values between the adversarial texts and the original texts. From the figure, we can see that at the cognitive level, our method requires minimal changes to the original texts, thus achieving recognition consistency and maintaining semantic consistency with the original texts in human reading comprehension.
\section{Conclusion}
This paper proposed a vision-fused attack (VFA) framework for generating powerful adversarial text. Our VFA uses the vision-merged solution space enhancement and perception-retained adversarial text selection strategy, producing more aggressive and stealthy adversarial text against NMT models. Extensive experiments demonstrated that VFA outperforms comparisons by significant margins both in attacking ability and imperceptibility enhancements. 
This paper indicates the important effect of multimodal correlations in current deep learning, which encourages future investigations on the corresponding topics, \emph{e.g.}, adversarial defense.

\section*{Acknowledgements}
This work was supported by Grant KZ46009501.

\section*{Contribution Statement}
Yanni Xue and Haojie Hao contributed equally to this work. Jiakai Wang is the corresponding author.

\bibliographystyle{named}

\newpage

\appendix

\section{Structure of Appendix\label{section:Structure}}

The appendix is structured into the following sections. We start by providing background information on adversarial text against machine translation models in section \ref{section:Background}. Subsequently, we present an overview of our proposed vision-fused attack, complete with pseudocode in section \ref{section:Overview}. In section \ref{section:Experiment}, we provide the specifics of our experiments, including datasets, metrics, models and detailed explanations of some experimental steps. The detailed implementation of compared methods and more examples are presented in section \ref{section:Compared} and section \ref{section:Case} respectively.

Additionally, a video showcasing query experiments on large language models (LLMs) is supplemented in our code repository. We randomly select three adversarial texts generated by our VFA and test them on LLMs such as ChatGPT and ERNIE, then compare them with other methods such as DRTT and Targeted Attack. The left half of the video shows the translation of the original text, while the right half shows the translation of the adversarial text. We first exhibit the attack effectiveness of VFA, and then exhibit the attack effectiveness of other methods. In the process of displaying the results, we provided a reference translation (Ground Truth) below the translation results for comparison of attack effects.

\section{Background\label{section:Background}}

In this section, we begin by presenting common definitions of adversarial text against machine translation models. Additionally, we suggest relevant works associated with human perception in NLP tasks. To illuminate the distinctions between our approach and previous works, we will delve into these details in the following subsections.

\subsection{Adversarial Text for NMT}

Drawing inspiration from the definition of adversarial text in the classification task, adversarial text meeting the criteria in machine translation should ensure similarity to the original sentence while causing the model to fail~\cite{ebrahimi2018adversarial}. In the context of machine translation, causing the model to fail corresponds to a degradation in the quality of the translated output. Consequently, the definition of adversarial text in machine translation is provided as follows:

\subsubsection{\textbf{Definition 1:}}

For a given source sentence $\mathbf{x}$, we generate an adversarial text $\mathbf{x_{\delta}}$, and employ a semantic similarity function $\operatorname{sim}(\mathbf{x},\mathbf{x_{\delta}})$ to evaluate sentence similarity, which typically uses $\operatorname{BLEU}$ in NMT tasks. Valid adversarial text must satisfy the following conditions:
\begin{equation}
\begin{aligned}
    \left\{
    \begin{aligned}
        & \operatorname{sim}(\mathbf{x}, \mathbf{x_{\delta}}) > \alpha_1, \\
        & \operatorname{sim}(\mathbf{y'}, \mathbf{y}) - \operatorname{sim}(\mathbf{y'_{\delta}}, \mathbf{y}) > \alpha_2.
    \end{aligned}
    \right.
\end{aligned}
\end{equation}
Here, $\mathbf{y'}$ and $\mathbf{y'_{\delta}}$ represent the translation results of the original sentence and the adversarial text, respectively, and $\mathbf{y}$ represents the target sentence. The parameter $\alpha_1$ is established as the lower limit, representing the lowest score of similarity between $\mathbf{x}$ and $\mathbf{x_{\delta}}$. The parameter $\alpha_2$ is also set as the lower limit, denoting the lowest decrease in translation quality.

\subsubsection{\textbf{Definition 2:}} 

According to definition 1, it has been pointed out that evaluating the similarity between $\mathbf{y}$ and $\mathbf{y'_{\delta}}$ poses a potential pitfall~\cite{zhang2021crafting}. If $\mathbf{x}$ has been changed to $\mathbf{x_{\delta}}$, the reference for $\mathbf{y'_{\delta}}$ should also be adjusted. Consequently, a double-round translation is introduced to filter valid adversarial text. The formulation for the similarity between the source sentence and adversarial text $\operatorname{d_{\text{src}}}(\mathbf{x}, \mathbf{x_\delta})$ is given by:
\begin{equation}
\begin{aligned}
    \operatorname{d_{\text{src}}}(\mathbf{x}, \mathbf{x_\delta}) = \frac{\operatorname{sim}(\mathbf{x}, \mathbf{\hat x}) - \operatorname{sim}(\mathbf{x_\delta}, \mathbf{\hat x_\delta})}{\operatorname{sim}(\mathbf{x}, \mathbf{\hat x})}> \alpha.
\end{aligned}
\end{equation}
Here, $\mathbf{\hat x}$ is reconstructed through two translation models: one is the target translation model, translating from language A to B, and the other is an auxiliary translation model, translating from language B to A. $\mathbf{\hat x_{\delta}}$ is also reconstructed, with $\alpha$ denoting the lowest degree of degradation in translation quality. Additionally, since division by $\operatorname{sim}(\mathbf{x}, \mathbf{\hat x})$ may encounter auxiliary errors, a further definition for valid adversarial text has been proposed in a recent work~\cite{lai2022generating}. The definition is as follows:

\begin{equation}
\begin{aligned}
    \left\{
    \begin{aligned}
        & \operatorname{d_{\text{src}}}(\mathbf{x}, \mathbf{x_\delta}) = \frac{\operatorname{sim}(\mathbf{x}, \mathbf{\hat x}) - \operatorname{sim}(\mathbf{x_\delta}, \mathbf{\hat x_\delta})}{\operatorname{sim}(\mathbf{x}, \mathbf{\hat x})}> \alpha, \\
        & \operatorname{d_{\text{tgt}}}(\mathbf{y}, \mathbf{y^{'}_\delta}) = \frac{\operatorname{sim}(\mathbf{y}, \mathbf{\hat y}) - \operatorname{sim}(\mathbf{y'_\delta}, \mathbf{\hat y'_\delta})}{\operatorname{sim}(\mathbf{y}, \mathbf{\hat y})} <\gamma.
    \end{aligned}
    \right.
\end{aligned}
\end{equation}
This way, evaluate not only the relative difference between the similarity of the source sentence $\mathbf{x}$ and its reconstruction through translation but also reconstruct the reference and adversary’s translation results to assess relative changes, thereby eliminating the effect of the auxiliary model where $\alpha$ and $\gamma$ are threshold values.

\subsubsection{\textbf{Differences between Ours and Previous Works}} 

The core of these approaches is to address the issue of insufficient reference translation, which narrows down the solution space, leading to a lower attack success rate. In our work, we introduce visual perturbations to generate adversarial text. This strategy effectively mitigates the problem of inadequate reference translation caused by semantic alterations in the adversarial text and introduces a new visual solution space, thereby enhancing the ASR. As the visual solution space closely aligns with human perception, our adversarial text also exhibits higher perceptual retention compared to other approaches. The definition of adversarial text differs accordingly.

\subsection{Perception Correlated Attacks in NLP}

In specific text classification tasks, some studies incorporate human perception to craft adversarial text. By making minor alterations at the character or word level, such as additions, deletions, or substitutions, these changes may go unnoticed by humans~\cite{ebrahimi2018hotflip}. Particularly in Latin languages, preserving the first and last letters of a word does not influence human perception~\cite{pelli2003remarkable}. Similarly, in the Chinese context, some studies utilize character structure to deconstruct characters into parts and randomly replace them with visually similar glyphs, rendering the changes imperceptible to humans~\cite{zhang2021argot}. Informed by these studies, we propose an expanded approach that involves a visually oriented search strategy constrained by perceptual function to evaluate the impact of alterations.

\section{Overview of Vision-fused Attack\label{section:Overview}}

The Vision-fused Attack algorithm, which is illustrated in Algorithm \ref{alg:algorithm}, aims to generate adversarial text by using auxiliary translation models, word importance models, and sentence similarity models. Given a source sentence $\textbf{x}$, a reference sentence $\textbf{y}$, a victim model $\mathbb{M}$, and tool models $\textbf{M}_{\text{aux}}$, $\textbf{M}_{\text{mlm}}$, and $\textbf{M}_{\text{sim}}$, the algorithm iterates over translated sentence segments $\hat{\textbf{x}}$ and applies perturbations based on word importance scores. These perturbations include radial and pixel modifications. The algorithm evaluates the perturbed sentences, considering both sentence and visual similarity thresholds ($\beta$ and $\theta$). The process aims to generate an adversarial text $\textbf{x}_\delta$ that, when passed through the victim model, maintains high perceptual similarity with the source sentence while degrading model performance. The output is the final generated adversarial text through the vision-fused attack.

\begin{algorithm}[htb]
    \caption{Vision-fused Attack}
    \label{alg:algorithm}
    \textbf{Input}: \par
        \hspace*{0.6cm}%
        \begin{minipage}{.8\textwidth}%
            $\textbf{x}$: source sentence \par
            $\textbf{y}$: reference sentence \par
            $\mathbb{M}$: victim model \par
            $\textbf{M}_{\text{aux}}$: auxiliary model \par
            $\textbf{M}_{\text{mlm}}$: model for word importance \par
            $\textbf{M}_{\text{sim}}$: model for sentence similarity \par
        \end{minipage}%

    \textbf{Parameter}: \par
        \hspace*{0.6cm}%
        \begin{minipage}{.8\textwidth}%
            $\beta, \theta$: sentence and visual similarity threshold \par
        \end{minipage}%
        
    \textbf{Output}: \par
        \hspace*{0.6cm}%
        \begin{minipage}{.8\textwidth}%
            $\textbf{x}_\delta$: adversarial text \par
        \end{minipage}%

    \textbf{Begin}: \par
    \begin{algorithmic}[1]
        \STATE $\textbf{x}_\delta \gets \textbf{x}$
        \STATE $\hat{\textbf{x}} \gets \textbf{M}_{\text{aux}}(\textbf{x})$ \hfill // Reverse Translation
        \FOR{$\hat{\textbf{x}}_{i}$ in $\hat{\textbf{x}}$}
            \IF{$\textbf{M}_{\text{sim}}(\hat{\textbf{x}}_{i}, \textbf{x})<\beta$}
                \STATE continue
            \ENDIF
            \STATE $\textbf{w}_{\text{imp}} \gets \operatorname{\textbf{M}_{\text{mlm}}}(\hat{\textbf{x}}_{i})$  \hfill // Semantic Importance Ranking
            \STATE $\textbf{S} \gets \textbf{S}_{\text{rad}} \cup \textbf{S}_{\text{pix}}$ \hfill // Text-Image Transformation
            \STATE $\hat{\textbf{x}}_{\delta}^i \gets \hat{\textbf{x}}_{i}$
            \FOR{$\textbf{w}_{j}$ in $\textbf{w}_{\text{imp}}$}
                \FOR{$c$ in $\textbf{w}_j$}
                    \IF{$\text{rep}(c_{\delta},c) == 1$}
                        \STATE $\text{Replace}(\hat{\textbf{x}}_{\delta}^i)$  \hfill // Improved Word Replacement
                    \ENDIF
                    \STATE $l \gets \mathbb{L}(\text{p}(\hat{\textbf{x}}_{\delta}^i),\text{p}(\textbf{x})) + \epsilon \cdot \mathbb{L}(\text{p}(c_{\delta}),\text{p}(c))$
                    \IF{$l<\theta$}
                        \STATE $\text{Back}(\hat{\textbf{x}}_{\delta}^i)$  \hfill // Global Perceptual Constraint
                        \STATE continue
                    \ENDIF
                \ENDFOR
            \ENDFOR
            \IF{$\textbf{M}_{\text{sim}}(\hat{\textbf{x}}_{\delta}^i, \textbf{x}) < \textbf{M}_{\text{sim}}(\textbf{x}_{\delta}, \textbf{x})$}
                \STATE $\textbf{x}_{\delta} \gets \hat{\textbf{x}}_{\delta}^i$  \hfill // Get Final Adversarial Text
            \ENDIF
        \ENDFOR
        \STATE \textbf{return} $\textbf{x}_{\delta}$
    \end{algorithmic}
    \textbf{End}
\end{algorithm}

\section{Details of Experiment\label{section:Experiment}}

\subsection{Statistics of Datasets}

For the datasets, we choose the validation set of WMT19, WMT18, and TED datasets for the Zh-En translation task and the test set of the ASPEC dataset for the Ja-En translation task. We use the name of the dataset to refer to the test dataset we are using. Here are some statistics, including the number of text pairs (ntp) as well as the average length of the source and target texts (als and alt), reported in Table \ref{tab:dataset}.

\begin{table}[htbp]
    \footnotesize
    \centering
    \tabcolsep=0.1cm
    \renewcommand{\arraystretch}{1}
    \begin{tabular*}{\hsize}{@{}@{\extracolsep{\fill}}ccccc@{}}
        \toprule
            Dataset & WMT19 & WMT18 & TED & ASPEC \\
        \midrule
            ntp & $3981$   & $2001$   & $1958$  & $1812$   \\
            als & $40.87$  & $41.96$  & $31.12$ & $46.59$  \\
            alt & $152.50$ & $143.22$ & $89.71$ & $141.85$ \\
        \bottomrule
    \end{tabular*}
    \caption{Statistics of datasets.}
    \label{tab:dataset}
\end{table}

\subsection{Calculation of Metrics}

We calculate the BLEU scores for the translated English texts to evaluate the translation quality. We employ the \textit{word\_tokenize} function from the nltk\footnote{\href{https://github.com/nltk/nltk}{https://github.com/nltk/nltk}} library to tokenize English texts and use the \textit{sentence\_bleu} function from the same library to compute BLEU scores, where \textit{SmoothingFunction.method1} is selected as the smoothing function.

We use the SSIM value to evaluate the perceptual similarity between the adversarial text and the original text. We first use the Pillow\footnote{\href{https://github.com/python-pillow/Pillow}{https://github.com/python-pillow/Pillow}} library to convert both the adversarial text and the original text into images using the "simsun" font with the font size 18. Then, we use the \textit{ssim} function in the scikit-image\footnote{\href{https://github.com/scikit-image/scikit-image}{https://github.com/scikit-image/scikit-image}} library to calculate the SSIM value for the two images.

\begin{table*}[htbp]
    \begin{center}
        \begin{minipage}{0.49\textwidth}
            \footnotesize
            \centering
            \tabcolsep=0.1cm
            \renewcommand{\arraystretch}{1}
            \begin{tabular*}{\hsize}{@{}@{\extracolsep{\fill}}ccccccc@{}}
                \toprule
                    Method & Ours & ADV & DRTT & HotFlip & Targeted & Seq2Sick   \\
                \midrule
                    option1  & 45     & 12     & 71     & 26      & 64     & 60       \\
                    option2  & 44     & 45     & 36     & 19      & 32     & 18       \\
                    option3  & 51     & 43     & 67     & 22      & 19     & 26       \\
                    option4  & 50     & 35     & 73     & 46      & 44     & 36       \\
                    option5  & 75     & 17     & 37     & 63      & 25     & 70       \\
                    option6  & 48     & 68     & 72     & 22      & 36     & 25       \\
                    option7  & 46     & 36     & 63     & 42      & 48     & 11       \\
                    option8  & 33     & 38     & 65     & 21      & 12     & 28       \\
                    option9  & 25     & 55     & 39     & 9       & 42     & 8        \\
                    option10 & 48     & 16     & 36     & 16      & 43     & 37       \\
                    option11 & 48     & 44     & 67     & 16      & 28     & 21       \\
                    option12 & 86     & 23     & 75     & 13      & 41     & 39       \\
                    option13 & 84     & 23     & 37     & 48      & 27     & 39       \\
                    option14 & 41     & 47     & 30     & 51      & 16     & 21       \\
                    option15 & 26     & 9      & 50     & 7       & 33     & 39       \\
                    option16 & 59     & 29     & 41     & 22      & 28     & 18       \\
                    option17 & 37     & 24     & 39     & 8       & 40     & 32       \\
                    option18 & 34     & 28     & 41     & 40      & 28     & 19       \\
                    option19 & 52     & 20     & 31     & 27      & 29     & 25       \\
                    option20 & 50     & 32     & 55     & 14      & 64     & 47       \\
                    score1{\color{red} ↑}&9.35&6.13&\textbf{9.76}&5.07&6.66&5.90   \\
                \bottomrule
            \end{tabular*}
            \caption{Statistics of "Semantic understanding" stage.}
            \label{tab:stage-1}
        \end{minipage}
        \hfill
        \begin{minipage}{0.49\textwidth}
            \footnotesize
            \centering
            \tabcolsep=0.1cm
            \renewcommand{\arraystretch}{1}
            \begin{tabular*}{\hsize}{@{}@{\extracolsep{\fill}}ccccccc@{}}
                \toprule
                    Method & Ours & ADV & DRTT & HotFlip & Targeted & Seq2Sick   \\
                \midrule
                    option1  & 43  & 11  & 41  & 21  & 43  & 38  \\
                    option2  & 43  & 43  & 24  & 18  & 29  & 15  \\
                    option3  & 48  & 40  & 27  & 18  & 13  & 26  \\
                    option4  & 46  & 30  & 37  & 45  & 28  & 26  \\
                    option5  & 69  & 15  & 20  & 62  & 24  & 41  \\
                    option6  & 43  & 66  & 35  & 13  & 18  & 19  \\
                    option7  & 44  & 35  & 46  & 40  & 28  & 10  \\
                    option8  & 32  & 36  & 36  & 20  & 9   & 18  \\
                    option9  & 17  & 55  & 18  & 8   & 26  & 7   \\
                    option10 & 45  & 16  & 20  & 14  & 36  & 19  \\
                    option11 & 43  & 42  & 44  & 14  & 18  & 12  \\
                    option12 & 81  & 22  & 41  & 10  & 33  & 14  \\
                    option13 & 79  & 18  & 23  & 38  & 17  & 34  \\
                    option14 & 41  & 44  & 17  & 30  & 9   & 15  \\
                    option15 & 16  & 9   & 22  & 6   & 18  & 19  \\
                    option16 & 54  & 28  & 21  & 17  & 17  & 8   \\
                    option17 & 33  & 23  & 22  & 8   & 22  & 27  \\
                    option18 & 30  & 26  & 22  & 36  & 20  & 5   \\
                    option19 & 51  & 19  & 16  & 23  & 23  & 17  \\
                    option20 & 48  & 28  & 44  & 12  & 19  & 37  \\
                    score2{\color{red} ↑}&\textbf{8.63}&5.77&5.49&4.31&4.29&3.88 \\
                \bottomrule
            \end{tabular*}
            \caption{Statistics of "Semantic comparison" stage.}
            \label{tab:stage-2}
        \end{minipage}
    \end{center}
\end{table*}

\subsection{Details of Tool Models}

We use many tool models in Reverse Translation and Semantic Importance Ranking to complete the corresponding tasks. The details are as follows:

\begin{itemize}
    \item In Reverse Translation Block, we use \textit{sentence-transformers/all-minilm-l6-v2}\footnote{\href{https://huggingface.co/sentence-transformers/all-minilm-l6-v2}{https://huggingface.co/sentence-transformers/all-minilm-l6-v2}} model as the multilingual sentence model to evaluate the semantic similarity between the reverse translation $\mathbf{\hat x}$ and the original text $\mathbf{x}$. 
    \item In Semantic Importance Ranking, we use \textit{hfl/chinese-bert-wwm-ext}\footnote{\href{https://huggingface.co/hfl/chinese-bert-wwm-ext}{https://huggingface.co/hfl/chinese-bert-wwm-ext}} as the masked language model to predict word probability of occurrence.
\end{itemize}

\subsection{Details of Common NMT Models}

The pure-text NMT models for both translation tasks are implemented using HuggingFace's Marian Model architecture.

\begin{itemize}
    \item For Zh-En translation task, we use \textit{Helsinki-NLP/opus-mt-zh-en}\footnote{\href{https://huggingface.co/Helsinki-NLP/opus-mt-zh-en}{https://huggingface.co/Helsinki-NLP/opus-mt-zh-en}} model as target model, and \textit{Helsinki-NLP/opus-mt-en-zh}\footnote{\href{https://huggingface.co/Helsinki-NLP/opus-mt-en-zh}{https://huggingface.co/Helsinki-NLP/opus-mt-en-zh}} model as auxiliary model.
    \item For Ja-En translation task, we use \textit{Helsinki-NLP/opus-mt-ja-en}\footnote{\href{https://huggingface.co/Helsinki-NLP/opus-mt-ja-en}{https://huggingface.co/Helsinki-NLP/opus-mt-ja-en}} model as target model, and \textit{Helsinki-NLP/opus-tatoeba-en-ja}\footnote{\href{https://huggingface.co/Helsinki-NLP/opus-tatoeba-en-ja}{https://huggingface.co/Helsinki-NLP/opus-tatoeba-en-ja}} model as auxiliary model.
\end{itemize}

In the testing of pixel-based NMT models, we use Multilingual Visrep model\footnote{\href{https://github.com/esalesky/visrep/tree/multi}{https://github.com/esalesky/visrep/tree/multi}} as the target model and auxiliary model.

\subsection{Details of LLM Test}

\begin{CJK*}{UTF8}{gbsn}
In the process of testing on LLM, we use a unified prompt: ``将下面内容翻译成英文：" and concatenate it with the adversarial text as input for LLM. There are three scenarios for the output of LLM, and we handle them separately as follows:

\begin{itemize}
    \item The output of LLM is the translation of the adversarial text, which can be directly used as the translation result without the need for processing.
    \item The output of LLM includes not only the translation of adversarial text but also the translation of prompts, which often contain descriptions of "translate" or "translation". Therefore, we filter prompts based on token "translat" and remove translation results that contain translation of prompts.
    \item LLM did not perform the translation task correctly and performed normal tasks such as continuation or rewriting. At this time, its output contains Chinese characters, so we filter this part based on whether it contains Chinese characters.
\end{itemize}

\end{CJK*}

\subsection{Details of Human Study}

To evaluate the impact of adversarial texts generated by our VFA and compared methods on reading comprehension, we conduct a human perception study on SurveyPlus\footnote{\href{https://www.surveyplus.cn/}{https://www.surveyplus.cn/}}, which is one of the most commonly used crowdsourcing platforms. 

We selected 105 subjects with a bachelor's degree or above and aged between 18 and 49 years old for testing. In the testing process, we selected 20 texts that meet the definition of attack success for each method and created them into a questionnaire\footnote{\href{https://www.surveyplus.cn/lite/5500199647838208}{https://www.surveyplus.cn/lite/5500199647838208}}. The testing process are divided into two stages: "Semantic understanding" and "Semantic comparison". The scores for each method and each option at two stages are detailly reported in Table \ref{tab:stage-1} and Table \ref{tab:stage-2}.

\section{Details of Compared Methods\label{section:Compared}}

\subsection{HotFlip}

The HotFlip attack crafts adversarial text by iteratively flipping the most crucial tokens in a text sequence through gradients. The original method was designed for English classification tasks, and we re-implemented it to suit Chinese-English and Japanese-English translation tasks, referring to the implementation in the platform named OpenAttack\footnote{\href{https://github.com/thunlp/OpenAttack}{https://github.com/thunlp/OpenAttack}}.

\subsection{Targeted Attack}

Targeted Attack propose an optimization problem, including an adversarial loss term and a similarity term, and use gradient projection in the embedding space to craft an adversarial text against NMT model. 
In this paper, we use \textit{IDEA-CCNL/Wenzhong-GPT2-110M}\footnote{\href{https://huggingface.co/IDEA-CCNL/Wenzhong-GPT2-110M}{https://huggingface.co/IDEA-CCNL/Wenzhong-GPT2-110M}} and \textit{rinna/japanese-gpt2-small}\footnote{\href{https://huggingface.co/rinna/japanese-gpt2-small}{https://huggingface.co/rinna/japanese-gpt2-small}} as the language model to train the FC layer. We use the Adam optimizer with a learning rate of 0.02 to solve the optimization problem, and set hyperparameter $\alpha \in \{10, 4, 1\}$, which is consistent with the original paper.

\subsection{Seq2Sick}

Seq2Sick is a targeted attack against NMT models, aimed at ensuring that the translation result does not overlap with the words at each position of the reference translation. They propose a projected gradient method to address the issue of discrete input space, then adopt group lasso to enforce the sparsity of the distortion, and develop a regularization technique to further improve the success rate.
In this paper, we modify it to adapt to the NMT model of our Zh-En and Ja-En tasks. For the regularization parameter $\lambda$ which balances the distortion and attack success rate, we keep it consistent with the original paper and set it to 1.

\subsection{DRTT}

DRTT propose a new criterion for NMT adversarial texts based on Doubly Round-Trip Translation, which can ensure the texts that meet their proposed criterion are the authentic adversarial texts. In this paper, we referred to the code provided in the original paper and trained the MLM for the Zh-En task using the training set of the WMT19 dataset, and trained the MLM for the Ja-En task using the training set of the ASPEC dataset. For the attacking process, hyper-parameters $\beta$ is set to 0.01 and $\gamma$ is set to 0.5 for Zh-En and Ja-EN tasks which is consistent with the original paper.

\subsection{ADV}

ADV is one of the attack methods used to improve model robustness in RoCBert~\cite{su2022rocbert}. This attack method aims to generate visually similar adversarial text for Chinese. 

\section{More Examples in Case Study\label{section:Case}}

In this section, we present more adversarial texts generated by our VFA, then we analyze and demonstrate the solution space and imperceptibility of these texts.

\subsection{Enhanced Solution Space Analysis}

\begin{figure}[htbp]
        \centering
        \includegraphics[width=0.49\textwidth]{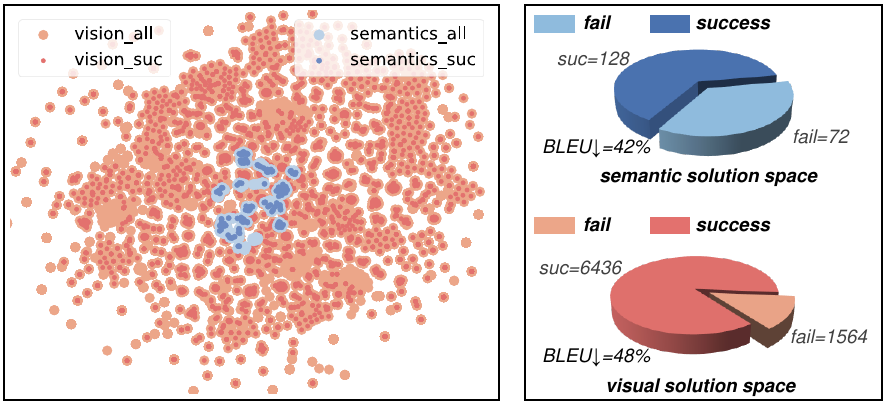}
        \caption{Input: ``他们千辛万苦才来到这儿。”.}
        \label{fig:supplementary-space-111}
    \end{figure}

\begin{figure}[htbp]
        \centering
        \includegraphics[width=0.49\textwidth]{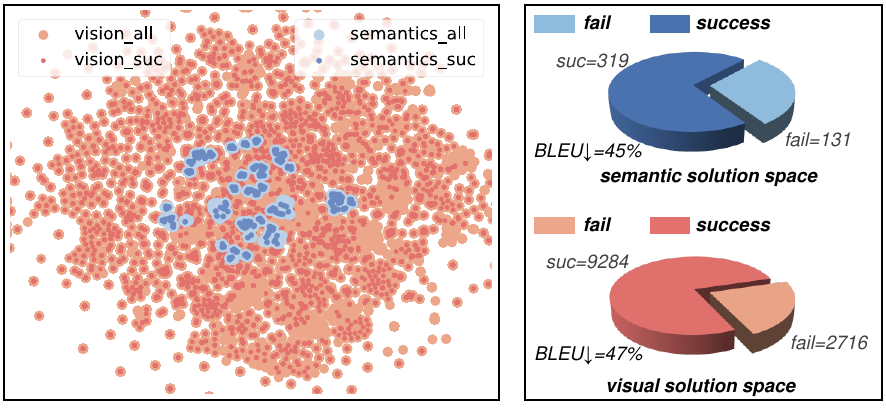}
        \caption{Input: ``第二所建造的房子是二号房屋。”.}
        \label{fig:supplementary-space-135}
    \end{figure}

\begin{figure*}[htbp]
        \begin{center}
            \begin{minipage}{0.49\textwidth}
                \centering
                \includegraphics[width=0.95\textwidth]{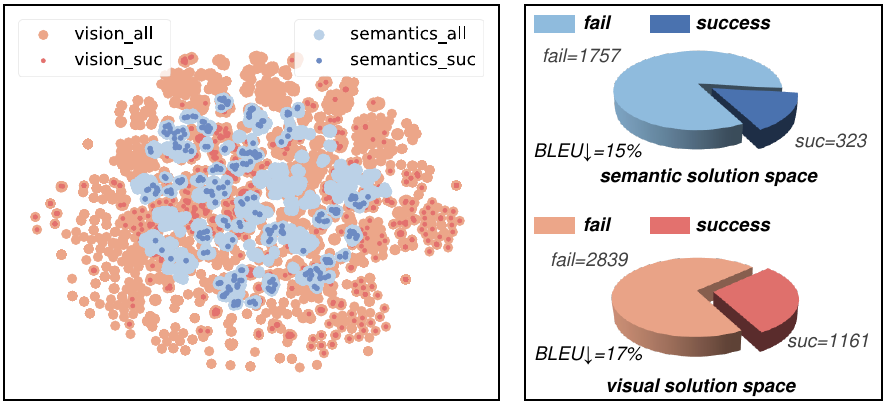}
                \caption{Input: ``我想,应该是雌青蛙产卵,雄青蛙使他们受精”.}
                \label{fig:supplementary-space-151}
            \end{minipage}
            \hfill
            \begin{minipage}{0.49\textwidth}
                \centering
                \includegraphics[width=0.95\textwidth]{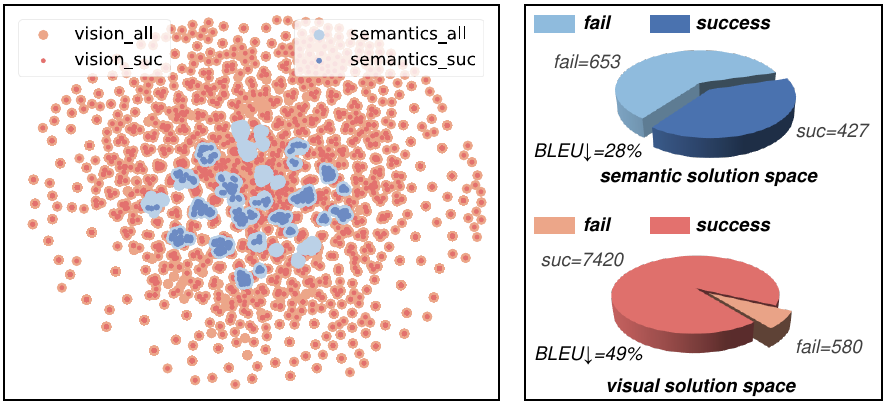}
                \caption{Input: ``她又继续,“那么狗狗呢？””.}
                \label{fig:supplementary-space-209}
            \end{minipage}
        \end{center}
    \end{figure*}

\begin{figure*}[htbp]
        \begin{center}
            \begin{minipage}{0.49\textwidth}
                \centering
                \includegraphics[width=0.95\textwidth]{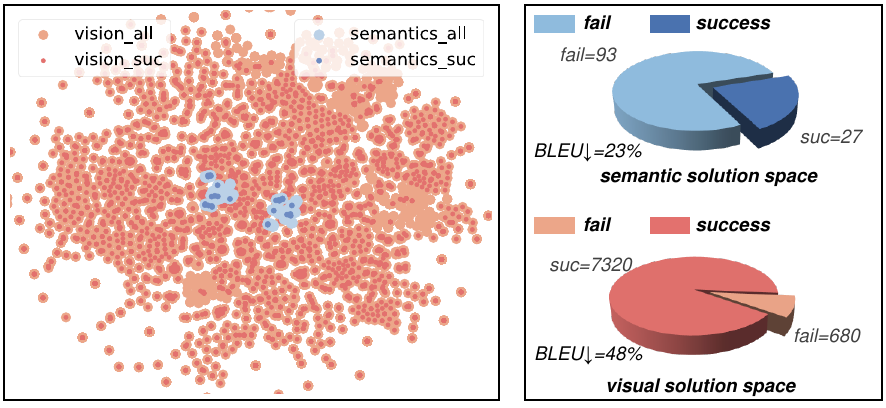}
                \caption{Input: ``你看见欧洲的更新非常频繁。”.}
                \label{fig:supplementary-space-256}
            \end{minipage}
            \hfill
            \begin{minipage}{0.49\textwidth}
                \centering
                \includegraphics[width=0.95\textwidth]{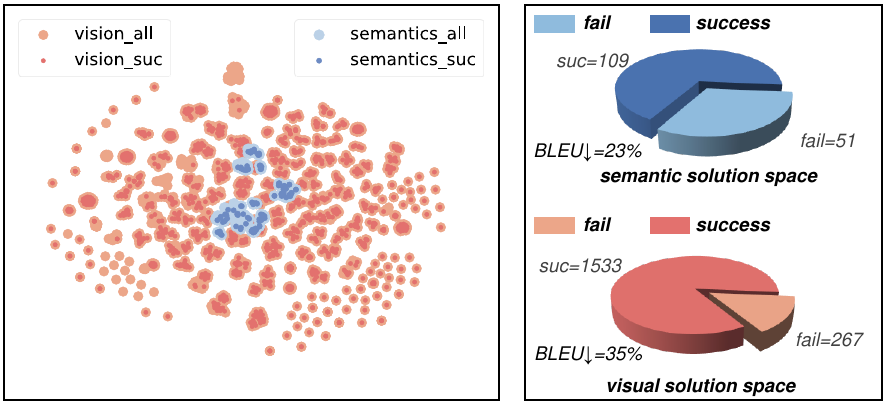}
                \caption{Input: ``这就是我们的成果。”.}
                \label{fig:supplementary-space-294}
            \end{minipage}
        \end{center}
    \end{figure*}

\begin{figure*}[htbp]
        \begin{center}
            \begin{minipage}{0.49\textwidth}
                \centering
                \includegraphics[width=0.95\textwidth]{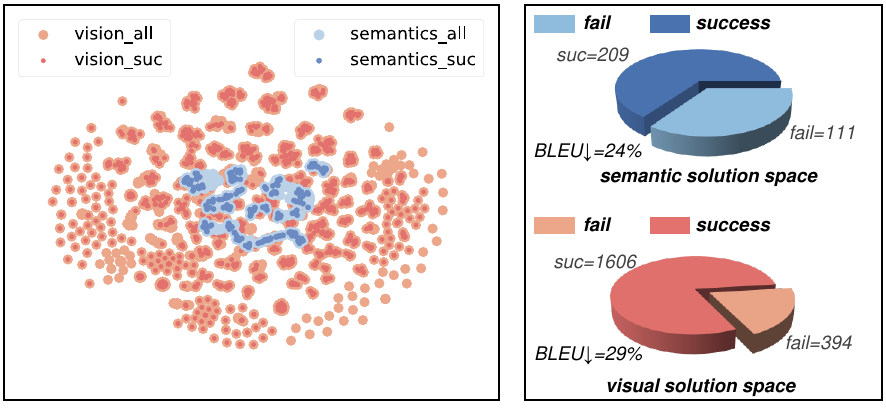}
                \caption{Input: ``那么我们必须做什么？”.}
                \label{fig:supplementary-space-512}
            \end{minipage}
            \hfill
            \begin{minipage}{0.49\textwidth}
                \centering
                \includegraphics[width=0.95\textwidth]{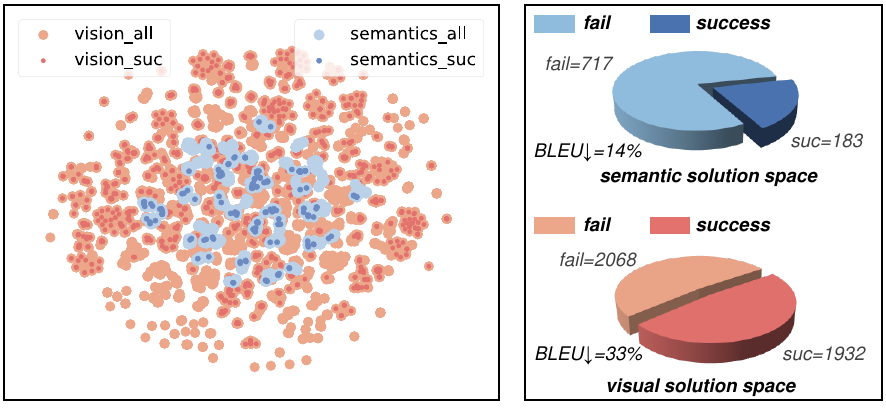}
                \caption{Input: ``我恳求你们参与推动的到底是什么？”.}
                \label{fig:supplementary-space-513}
            \end{minipage}
        \end{center}
    \end{figure*}

\begin{figure*}[htbp]
        \begin{center}
            \begin{minipage}{0.49\textwidth}
                \centering
                \includegraphics[width=0.95\textwidth]{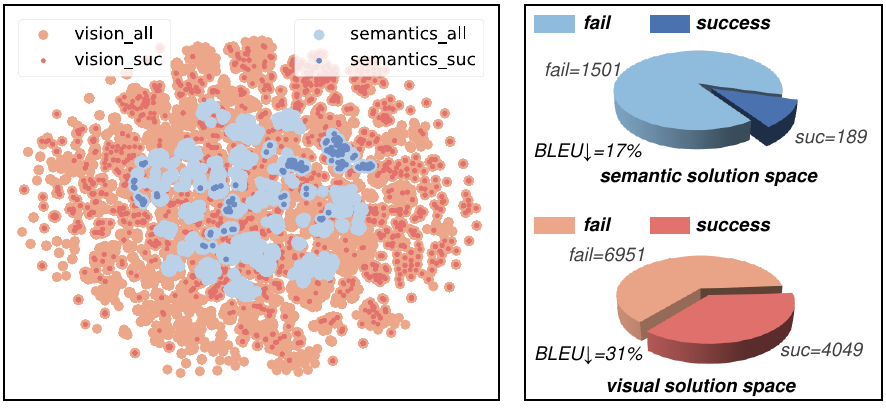}
                \caption{Input: ``放那儿烧个六十年，然后没了”.}
                \label{fig:supplementary-space-531}
            \end{minipage}
            \hfill
            \begin{minipage}{0.49\textwidth}
                \centering
                \includegraphics[width=0.95\textwidth]{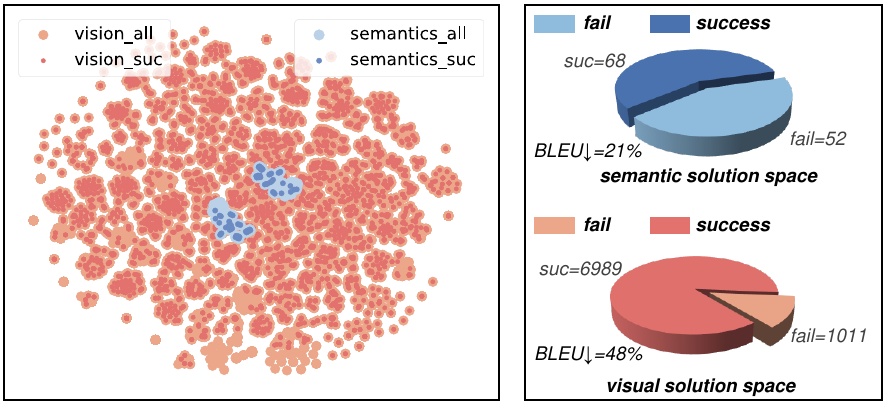}
                \caption{Input: ``经过思考以后，我想出了一个”.}
                \label{fig:supplementary-space-609}
            \end{minipage}
        \end{center}
    \end{figure*}

Figure \ref{fig:supplementary-space-111}-\ref{fig:supplementary-space-609} provides schematic representations of the solution space and statistical data for more examples. From the solution space figures of these examples, it can be witnessed that the scope of the enhanced solution space is broader than that of the semantic solution space. In Addition, the pie charts show that the enhanced solution space generates more aggressive texts than the semantic solution space, which proven by the BLEU decrease of the two approaches. More examples indicate the result that the enhanced solution space can obtain more diverse adversarial texts, providing more possibilities to find more effective adversarial texts, and enhanced solution space helps generate more texts with stronger aggressiveness.

\subsection{Imperceptibility Analysis}

\begin{figure*}[htbp]
        \begin{center}
            \begin{minipage}{0.49\textwidth}
                \centering
                \includegraphics[width=0.95\textwidth]{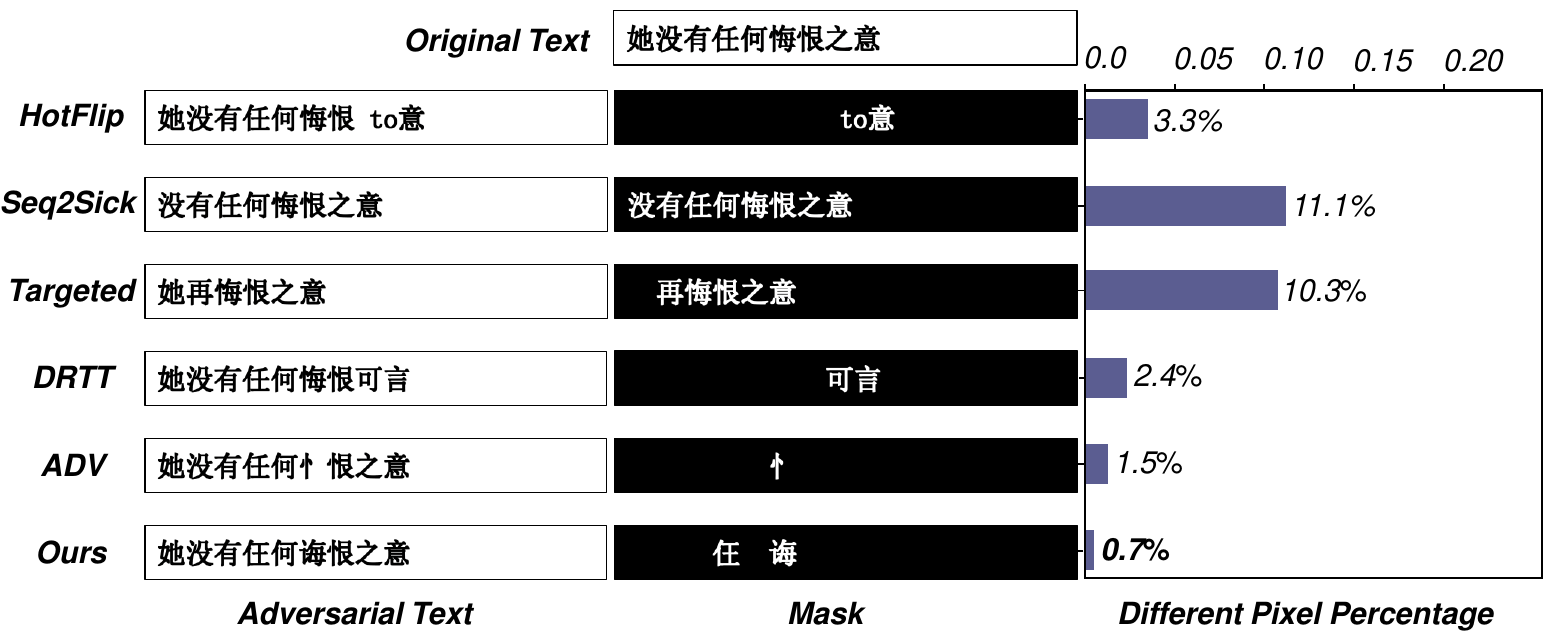}
                \caption{Input: ``她没有任何悔恨之意”.}
                \label{fig:supplementary-imperceptibility-1}
            \end{minipage}
            \hfill
            \begin{minipage}{0.49\textwidth}
                \centering
                \includegraphics[width=0.95\textwidth]{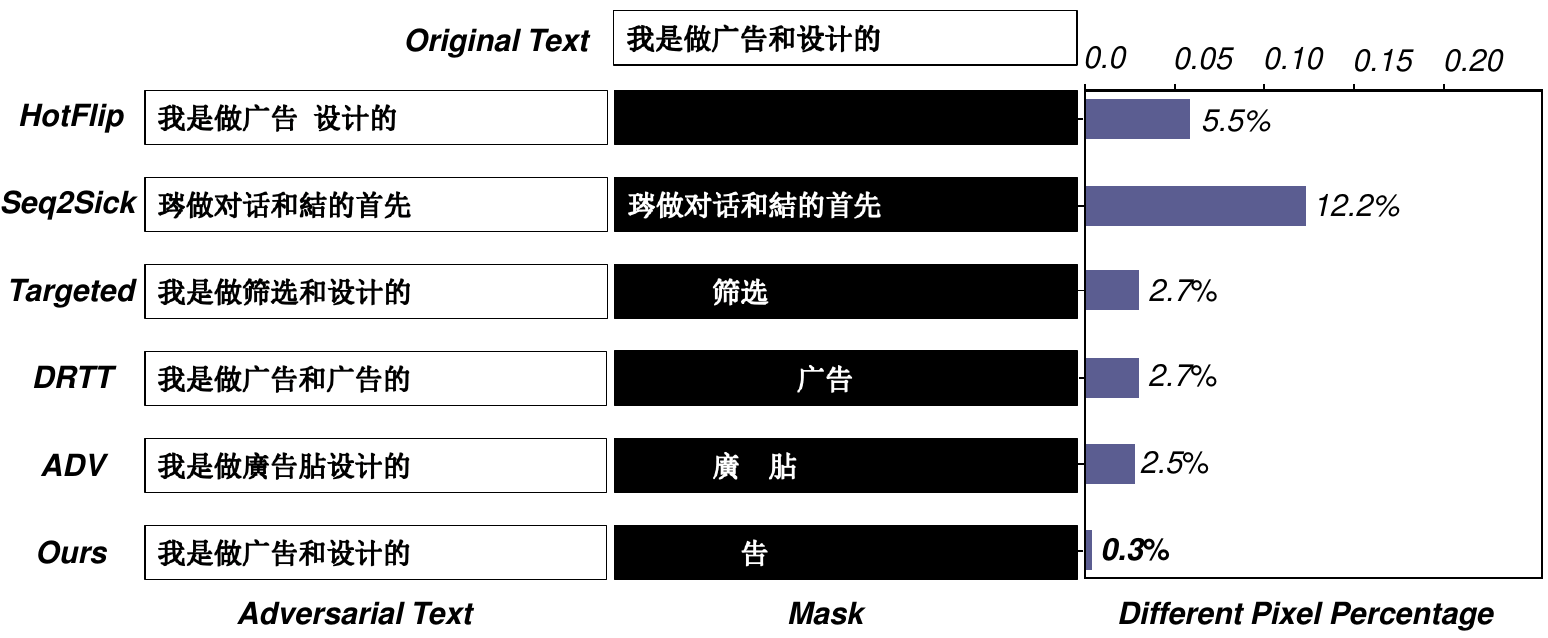}
                \caption{Input: ``我是做广告和设计的”.}
                \label{fig:supplementary-imperceptibility-2}
            \end{minipage}
        \end{center}
    \end{figure*}

\begin{figure*}[htbp]
        \begin{center}
            \begin{minipage}{0.49\textwidth}
                \centering
                \includegraphics[width=0.95\textwidth]{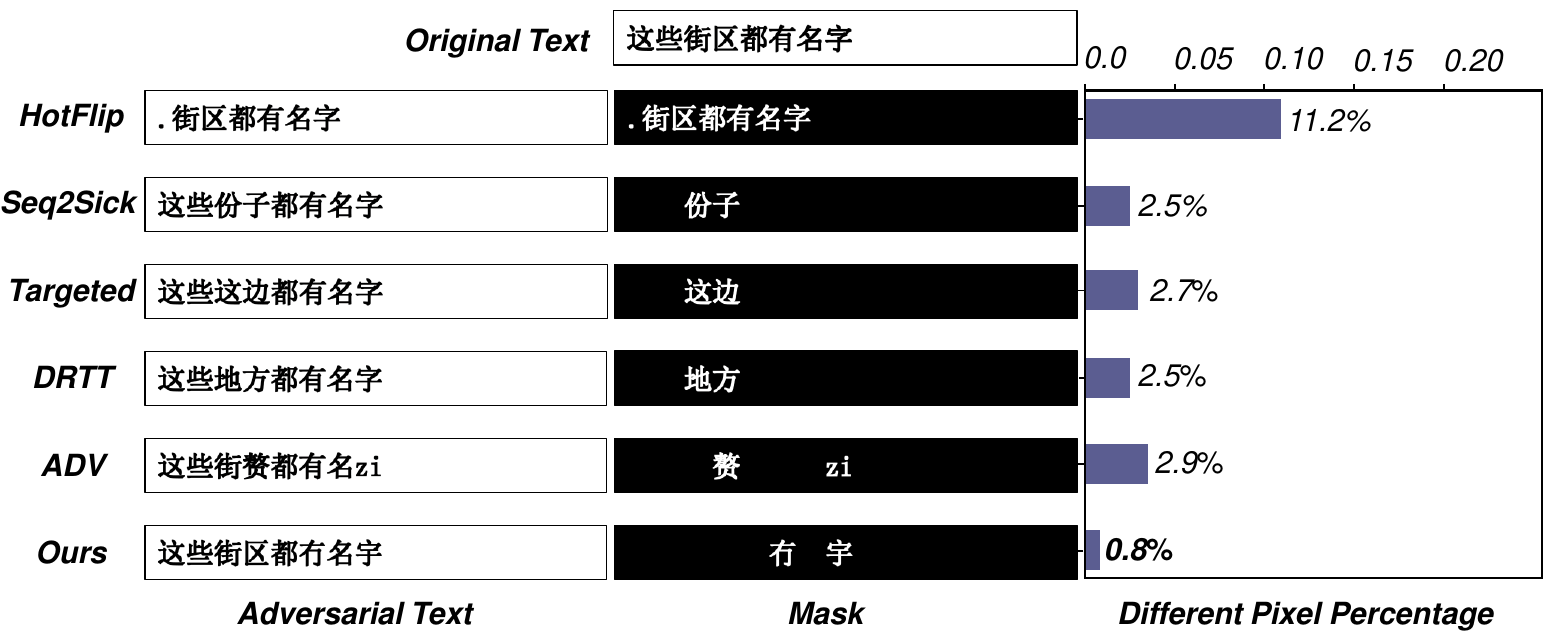}
                \caption{Input: ``这些街区都有名字”.}
                \label{fig:supplementary-imperceptibility-3}
            \end{minipage}
            \hfill
            \begin{minipage}{0.49\textwidth}
                \centering
                \includegraphics[width=0.95\textwidth]{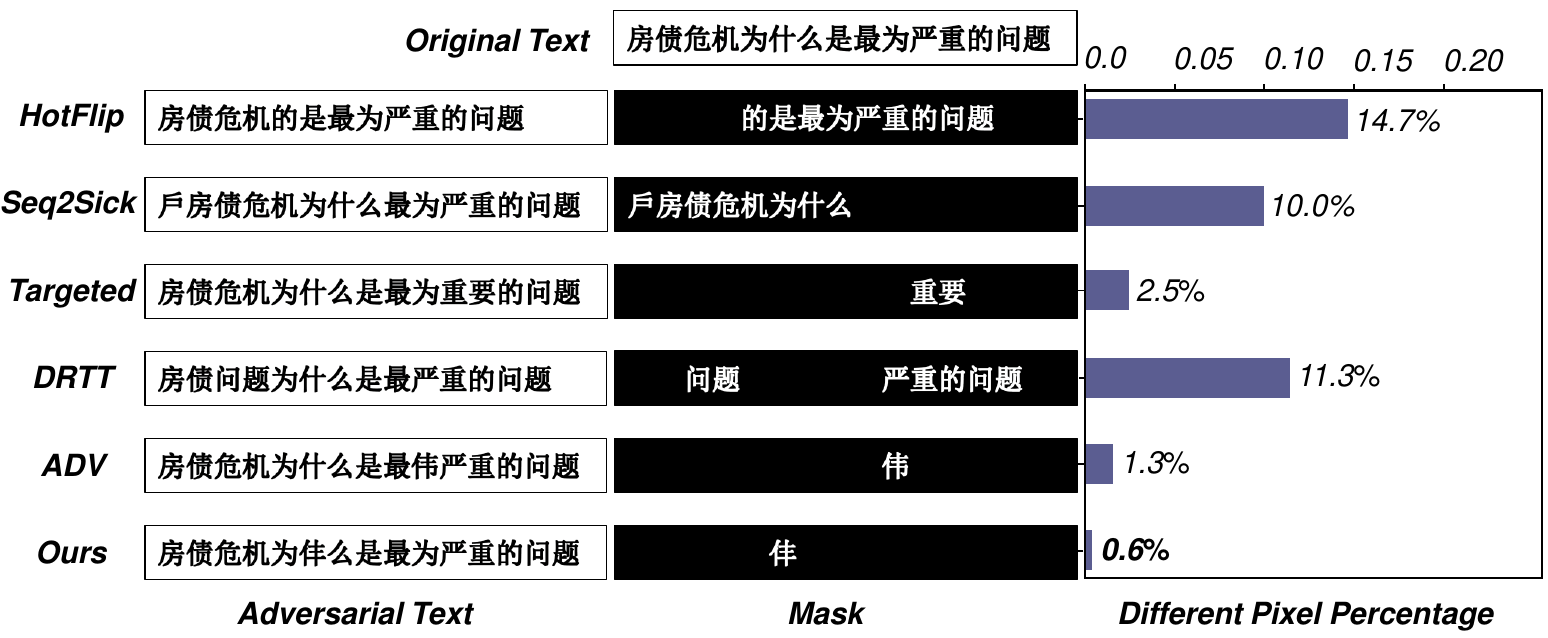}
                \caption{Input: ``房债危机为什么是最为严重的问题”.}
                \label{fig:supplementary-imperceptibility-4}
            \end{minipage}
        \end{center}
    \end{figure*}

\begin{figure*}[htbp]
        \begin{center}
            \begin{minipage}{0.49\textwidth}
                \centering
                \includegraphics[width=0.95\textwidth]{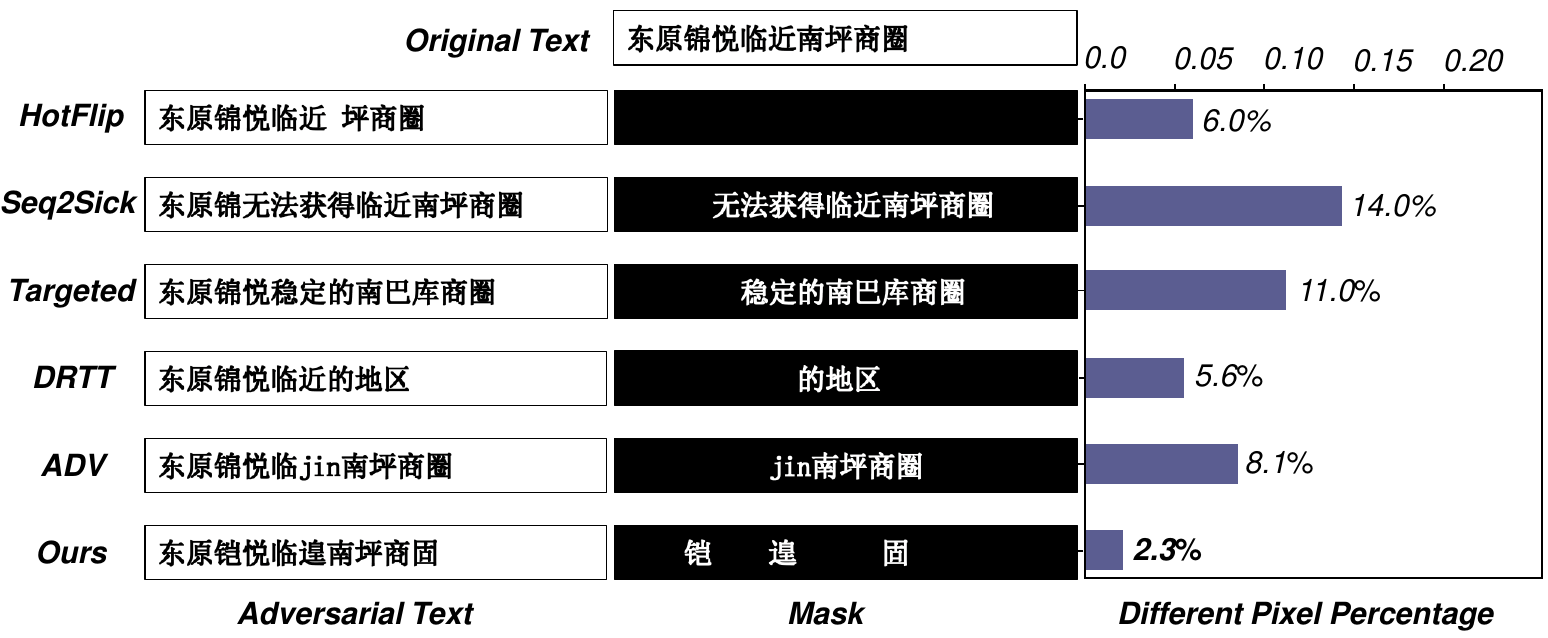}
                \caption{Input: ``东原锦悦临近南坪商圈”.}
                \label{fig:supplementary-imperceptibility-5}
            \end{minipage}
            \hfill
            \begin{minipage}{0.49\textwidth}
                \centering
                \includegraphics[width=0.95\textwidth]{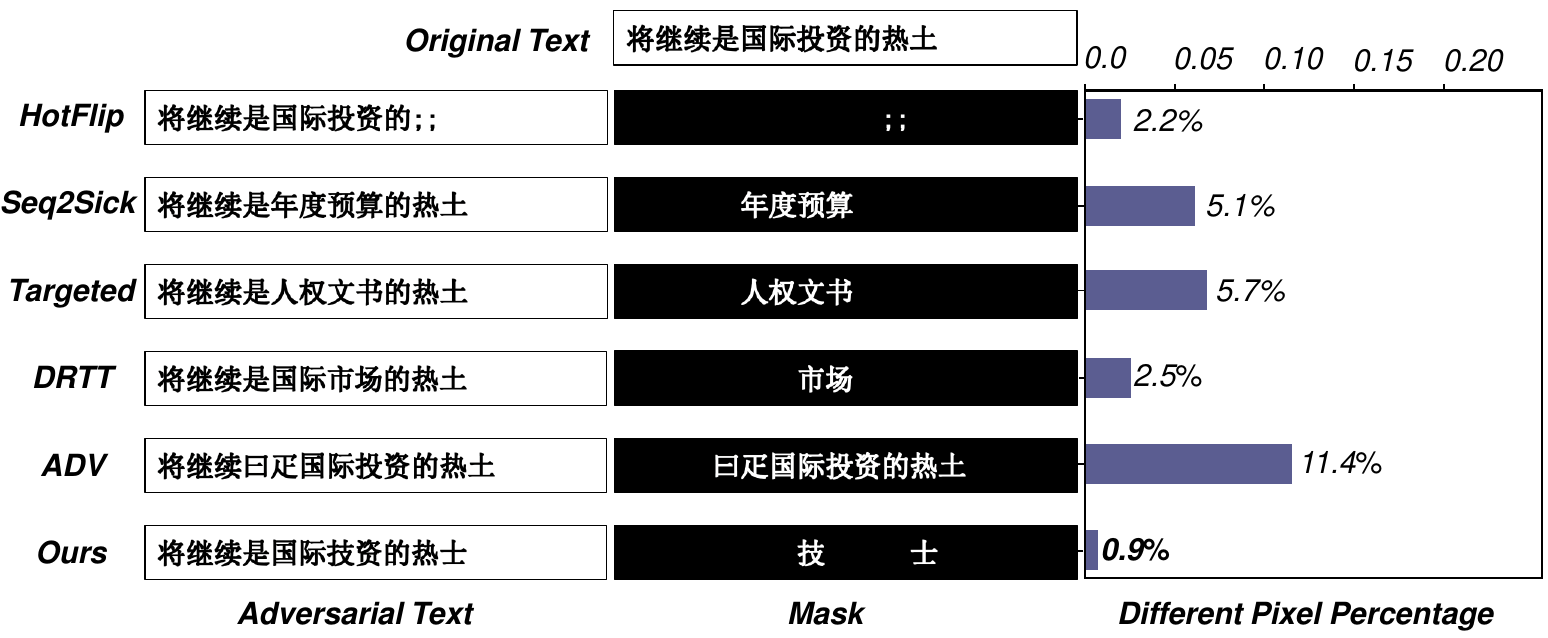}
                \caption{Input: ``将继续是国际投资的热土”.}
                \label{fig:supplementary-imperceptibility-6}
            \end{minipage}
        \end{center}
    \end{figure*}

\begin{figure*}[htbp]
        \begin{center}
            \begin{minipage}{0.49\textwidth}
                \centering
                \includegraphics[width=0.95\textwidth]{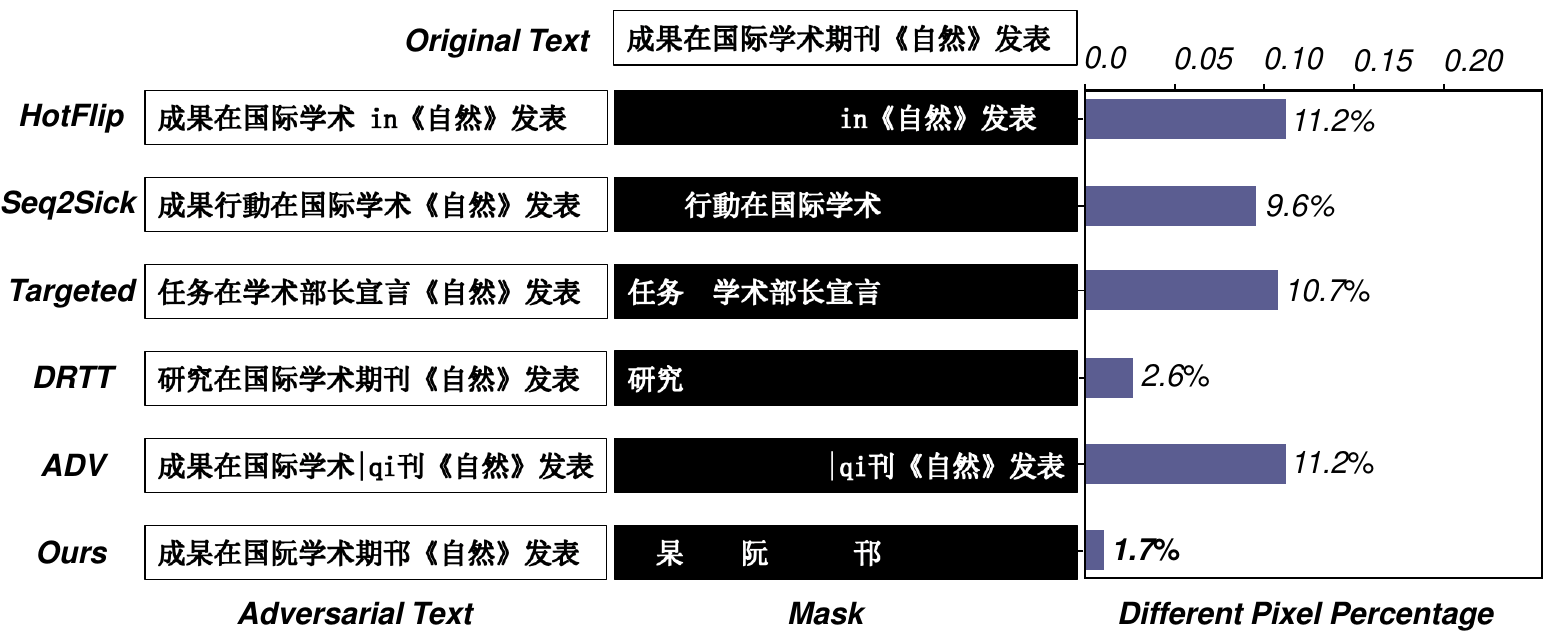}
                \caption{Input: ``成果在国际学术期刊《自然》发表”.}
                \label{fig:supplementary-imperceptibility-7}
            \end{minipage}
            \hfill
            \begin{minipage}{0.49\textwidth}
                \centering
                \includegraphics[width=0.95\textwidth]{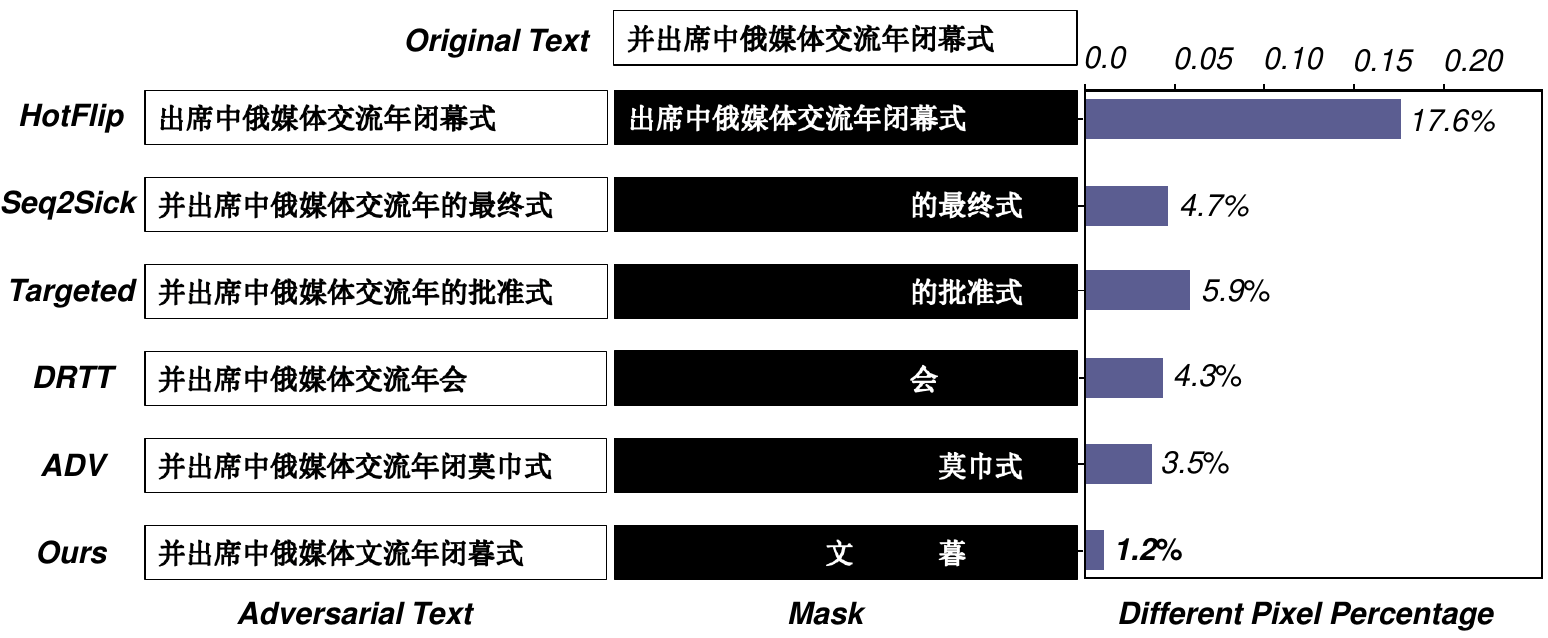}
                \caption{Input: ``并出席中俄媒体交流年闭幕式”.}
                \label{fig:supplementary-imperceptibility-8}
            \end{minipage}
        \end{center}
    \end{figure*}

\begin{figure*}[htbp]
        \begin{center}
            \begin{minipage}{0.49\textwidth}
                \centering
                \includegraphics[width=0.95\textwidth]{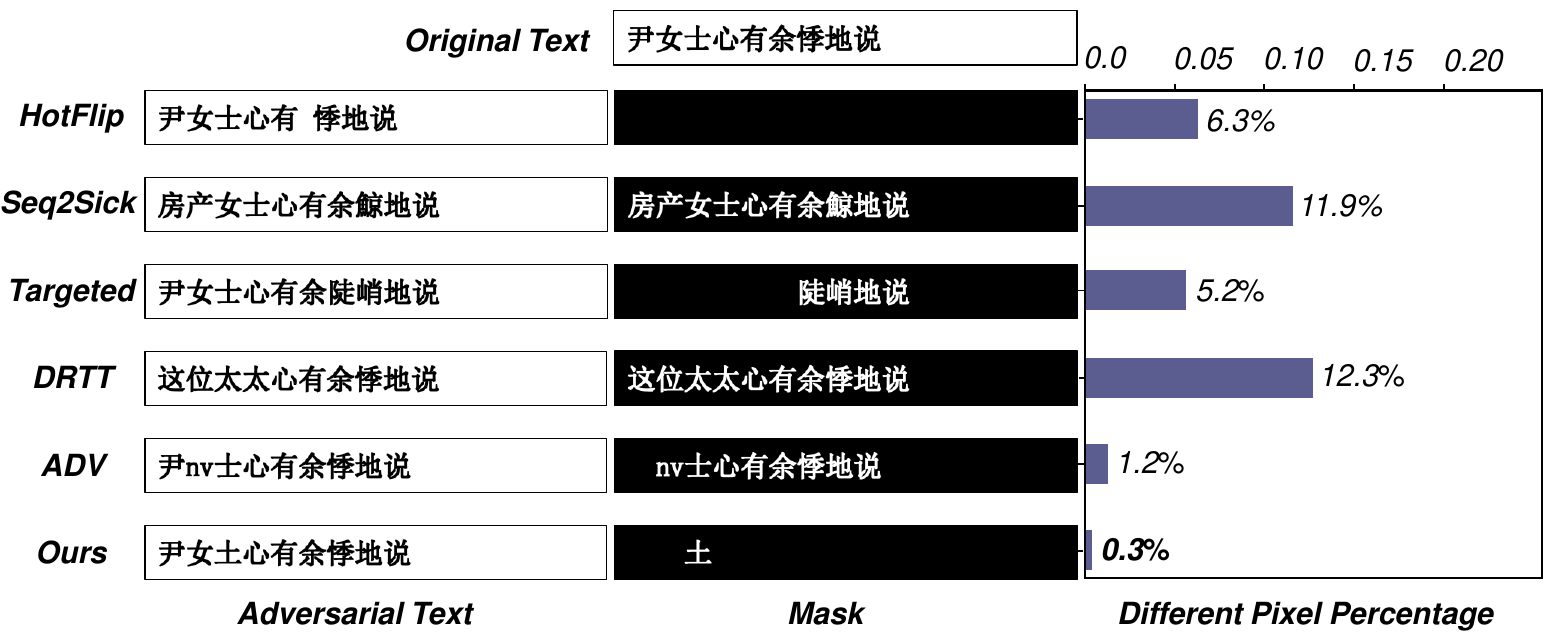}
                \caption{Input: ``尹女士心有余悸地说”.}
                \label{fig:supplementary-imperceptibility-9}
            \end{minipage}
            \hfill
            \begin{minipage}{0.49\textwidth}
                \centering
                \includegraphics[width=0.95\textwidth]{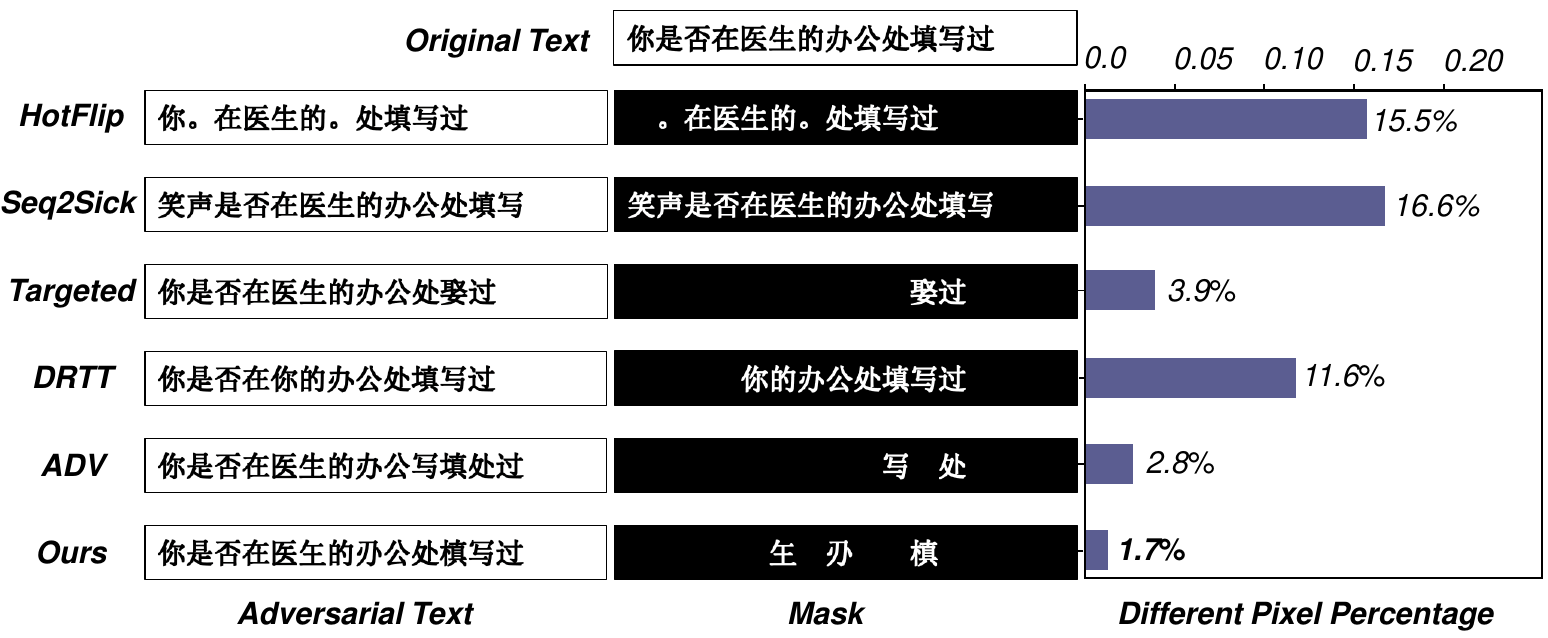}
                \caption{Input: ``你是否在医生的办公处填写过”.}
                \label{fig:supplementary-imperceptibility-10}
            \end{minipage}
        \end{center}
    \end{figure*}

Figure \ref{fig:supplementary-imperceptibility-1}-\ref{fig:supplementary-imperceptibility-10} provides schematic representations of the imperceptibility for more examples. From these examples, it can be witnessed that our method requires minimal changes to the original texts at the cognitive level, thus achieving recognition consistency and maintaining semantic consistency with the original texts in human reading comprehension.

\begin{CJK*}{UTF8}{gbsn}
\end{CJK*}

\begin{CJK*}{UTF8}{gbsn}
\end{CJK*}

\begin{CJK*}{UTF8}{gbsn}

\end{CJK*}

\begin{CJK*}{UTF8}{gbsn}

\end{CJK*}

\end{document}